\documentclass[11pt,a4paper]{article}
\usepackage{times,latexsym}
\usepackage{url}
\usepackage[T1]{fontenc}
\usepackage{amsmath}
\usepackage{amssymb}
\usepackage{nameref}
\usepackage{bm}
\usepackage{caption}
\usepackage{subcaption}

    \usepackage[acceptedWithA]{tacl2021v1}

\usepackage{tacl2021v1}

\usepackage{array}
\usepackage{graphicx}
\usepackage{float}
\usepackage{enumitem}
\usepackage{booktabs}
\usepackage{xcolor}
\usepackage{algorithm}
\usepackage{algorithmic}
\usepackage{cleveref}
\usepackage{xurl}

\definecolor{todo}{rgb}{1,0.5,0}
\definecolor{torevise}{rgb}{1,0.5,1}

\newcommand{\prompttemplate}[1]{%
  \noindent\fcolorbox{black!60}{gray!10}{%
    \parbox{\dimexpr\linewidth-2\fboxsep-2\fboxrule}{%
      \footnotesize
      #1
    }%
  }%
}

\newcolumntype{P}[1]{>{\centering\arraybackslash}m{#1}}

\usepackage{xspace,mfirstuc,tabulary}

\definecolor{greencolor}{rgb}{0.0, 0.5, 0.0}
\definecolor{redcolor}{rgb}{0.8, 0.0, 0.0}

\newif\iftaclinstructions
\taclinstructionsfalse %
\iftaclinstructions
\renewcommand{\confidential}{}
\renewcommand{\anonsubtext}{(No author info supplied here, for consistency with
TACL-submission anonymization requirements)}
\newcommand{\instr}
\fi

\iftaclpubformat %

\else

\fi

\title{Automatic Reviewers Fail to Detect Faulty Reasoning in Research Papers: A New Counterfactual Evaluation Framework}

\author{
    Nils Dycke
    \and
    Iryna Gurevych
   \\
   \ \\
   UKP Lab, Department of Computer Science and \\
   National Research Center for Applied Cybersecurity ATHENE \\
   Technical University of Darmstadt \\
   \url{www.ukp.tu-darmstadt.de}\\
 }

\date{}

\begin{document} %
\maketitle
\begin{abstract}
Large Language Models (LLMs) have great potential to accelerate and support scholarly peer review and are increasingly used as fully automatic review generators (ARGs). However, potential biases and systematic errors may pose significant risks to scientific integrity; understanding the specific capabilities and limitations of state-of-the-art ARGs is essential. We focus on a core reviewing skill that underpins high-quality peer review: detecting faulty research logic. This involves evaluating the internal consistency between a paper’s results, interpretations, and claims. We present a fully automated counterfactual evaluation framework that isolates and tests this skill under controlled conditions. Testing a range of ARG approaches, we find that, contrary to expectation, flaws in research logic have no significant effect on their output reviews. Based on our findings, we derive three actionable recommendations for future work and release our counterfactual dataset and evaluation framework publicly.\footnote{\url{https://github.com/UKPLab/tacl2026-counter-review-logic}}
\end{abstract}

\section{Introduction}

Scholarly peer review is the cornerstone of academic quality control \cite{birukou2011peerreview,bornmann_scientific_2011}. In this process, experts evaluate a submitted paper, writing \textit{review reports} that assess its soundness, clarity, and novelty \cite{jefferson2002prquality}. However, peer review is time-consuming and requires expertise \cite{waltman2023improve}. The growing volume of submissions, especially in AI research \cite{sculley2018avoiding}, and the shortage of qualified reviewers \cite{mccook2006peer} exacerbate the burden on reviewers.
The rise of Large Language Models (LLMs) like ChatGPT \cite{openai2024gpt4technicalreport} has led to an upsurge in LLM-assisted, and sometimes fully generated, review reports \cite{pmlr-v235-liang24b}, their official integration into peer review systems \cite{aaai2025aiPeerReview}, and growing research interest in automatic review generators (ARGs) \cite{yu-etal-2024-automated,d2024marg,yuan2022can}, machine-based scientific discovery \cite{weng2024cycleresearcher,lou2024aaar}, and peer review assistance \cite{chamoun-etal-2024-automated,dycke-etal-2023-nlpeer}. Yet, LLMs are prone to factual errors \cite{ji2023hallucination} and biases \cite{gallegos-etal-2024-bias}, making their unsupervised use as ARGs a potential threat to scientific integrity. It is imperative to empirically assess the capabilities of current ARGs to inform policymakers, guide responsible integration of LLMs into peer review, and clarify which aspects of reviewing require human judgment.

Current research on ARG performance yields highly inconsistent results with some studies reporting that ARGs show human-level or superior performance \cite{tyser2024ai,liang2024can,idahl2024openreviewer,kirtani2025revieweval} and commendable paper limitation identification abilities \cite{zhang2025reviewing,liu2023reviewergpt}, while others find automatic reviews to be generic \cite{du-etal-2024-llms} or failing to spot obvious errors \cite{son2025ai,li2025aspect}.
This variance stems from two key factors:  First, definitions of review quality differ widely. Second, peer review involves a diverse set of complex subtasks, such as recalling literature, reading comprehension, and step-wise reasoning \cite{dycke2025strictastructuredreasoningcritical}. Many of these depend on the paper's publication context such as related work, research trends, and evaluation standards; for instance, the assessment of state-of-the-art results is highly context-dependent since it is linked to the state-of-the-art at the time of reviewing.
Existing evaluations conflate these skills and frequently disregard the original reviewing context \cite{kuznetsov2024can}. While measuring overall reviewing performance is important, these factors confound existing evaluations inducing high variance across studies and prohibiting definitive conclusions about current ARG capabilities.

In this paper, to address these challenges, we evaluate ARGs in a controlled experiment  considering a skill that is prerequisite to reliable peer review and, as we show in \Cref{sec:rl}, depends on the paper content regardless of the reviewing context. Specifically, we test ARGs ability to identify \textit{faulty research logic} during reviewing. A paper's research logic encompasses its experimental design, the reasoning from measurements to interpretations, and derived findings. Assessing the soundness of a paper requires careful scrutiny of these elements and their logical relations.

We first propose a new model of paper soundness formalizing it as a research logic graph. Based on this, we introduce a new \textit{counterfactual evaluation} framework \cite{molnar2025,wu-etal-2021-polyjuice} that extracts the research logic from sound papers, introduces targeted misalignments through surgical edits, and compares reviews of original versus counterfactual versions to determine whether faulty research logic significantly impacts ARGs reviewing behavior.
Unlike prior error sensitivity analyses \cite{liu2023reviewergpt,zhang2025reviewing,li2025aspect,son2025ai}, our analysis eliminates confounding effects by using counterfactuals that intervene solely on the soundness of the research logic, and does not depend on an absolute notion of review quality by focusing on relative \textit{differences} between reviews. We develop and validate an LLM-based pipeline to produce counterfactual versions of  research papers with intentionally flawed research logic. The resulting dataset supports multiple downstream applications, including evaluation, explanation, and training data augmentation \cite{wu-etal-2021-polyjuice}. Our fully automatic approach is independent of human review data allowing to test ARGs on new and unseen papers at scale without risks of contamination \cite{sainz-etal-2023-nlp}. Our framework is general and applicable to any ARG regardless of its architecture.

Our experiments show that faulty research logic has no significant effect on the generated reviews of state-of-the-art ARGs raising serious concerns about their practical use during peer review. We analyze their automatic reviewing behavior and derive actionable recommendations to move the field forward including task design, human-LLM collaboration, and improved evaluation practices.
Overall, we contribute
\begin{itemize}[leftmargin=*]
    \item A formal model of paper soundness as a product of its underlying research design and reasoning, i.e., its \textit{research logic}.
    \item The first fully automatic counterfactual evaluation framework for ARGs focused on the detection of flawed research logic.
    \item A novel dataset of counterfactual research papers based on recent AI and NLP publications from major conferences including ACL, EMNLP, NeurIPS, and ICLR.
    \item Insights into the capabilities and limitations of state-of-the-art ARGs in identifying flawed research logic during review.
\end{itemize}

\section{Related Work}

\paragraph{Counterfactual Evaluation}

Counterfactuals are the key to study causal relationships \cite{pearl2000models}. By asking, \textit{what would a model predict under the same conditions if the input were changed?}, counterfactuals have become a standard approach in NLP for explaining  \cite{molnar2025,wang-etal-2024-survey} and evaluating models \cite{wu-etal-2021-polyjuice}. Prior work is largely limited to short texts \cite{wang-etal-2024-survey} or fixed narrative structures \cite{mu-li-2024-causal,wang2024beyond}. In contrast, we propose a pipeline to construct counterfactuals that intervene on full research papers. Further, while most approaches are designed for classification problems \cite{molnar2025}, we develop a framework tailored to full review reports as an output.

\paragraph{ARG Error Sensitivity Analysis}
Recent works on ARG evaluation \cite[][i.a.]{shin2025mind,du-etal-2024-llms,xu2024benchmarking} heavily rely on often noisy human review data as ground truth. To reduce this dependence, several works perform sensitivity analysis by introducing errors into papers. \citet{liu2023reviewergpt} manually inject errors into short scientific excerpts and inspect whether automatic reviews mention them. While their study is limited to $13$ scientific essays, we consider nearly $150$ full papers. \citet{zhang2025reviewing} and \citet{son2025ai} use retracted papers, leveraging self-reported errors to evaluate ARGs via LLM-based judges. These approaches depend on public retraction data and often involve issues, like plagiarized figures, that require external knowledge. In contrast, we focus on verifying research logic, which is self-contained within the paper. \citet{li2025aspect} automatically perturb paper sections (e.g. omitting implementation details) but unlike our study they exclusively consider automatic review scores without reports. \citet{tyser2024ai} propose a semi-automatic method to introduce broad error types (e.g. omitting related work) and check if they are detected. Our approach differs by introducing precise, targeted edits that affect only the paper's soundness. Finally, \citet{dycke2025strictastructuredreasoningcritical} use counterfactuals to investigate reasoning during peer review but focus on humans instead of ARGs.

In summary, this work is the first to isolate and study ARG's reasoning abilities during review generation in a fully automatic evaluation pipeline. 

\section{Research Logic} \label{sec:rl}
\newcommand{\pap}{p}
\newcommand{\fin}{F}
\newcommand{\con}{C}
\newcommand{\res}{R}
\newcommand{\met}{m}
\newcommand{\issound}{\text{snd}}
\newcommand{\entails}{\rightarrow}
\newcommand{\lland}[2]{\bigwedge\limits_{#1} #2}
\newcommand{\cfrl}{\text{CF}_\text{CR}}
\newcommand{\cfsn}{\text{CF}_\text{NE}}
\newcommand{\revrl}{\text{REV}_\text{CR}}
\newcommand{\revsn}{\text{REV}_\text{NE}}
\newcommand{\deltasn}{\Delta_\text{NE}}
\newcommand{\deltarl}{\Delta_\text{CR}}
\newcommand{\aterl}{\text{ATE}_\text{CR}}
\newcommand{\atesn}{\text{ATE}_\text{NE}}
\newcommand{\ate}{\text{ATE}}
Soundness, i.e. the correctness of a paper’s underlying scientific process, is a central criterion in peer review \cite{jefferson2002prquality}. To evaluate whether ARGs can identify flawed research design and reasoning, we first define soundness formally. We focus on AI and NLP domains as they are most common in prior work \cite{kuznetsov2024can} allowing direct comparison; unless stated otherwise, we refer to papers\footnote{We use the terms paper and manuscript interchangeably.} from only these fields.

\begin{figure}[t] 
  \centering
  \includegraphics[width=0.95\linewidth]{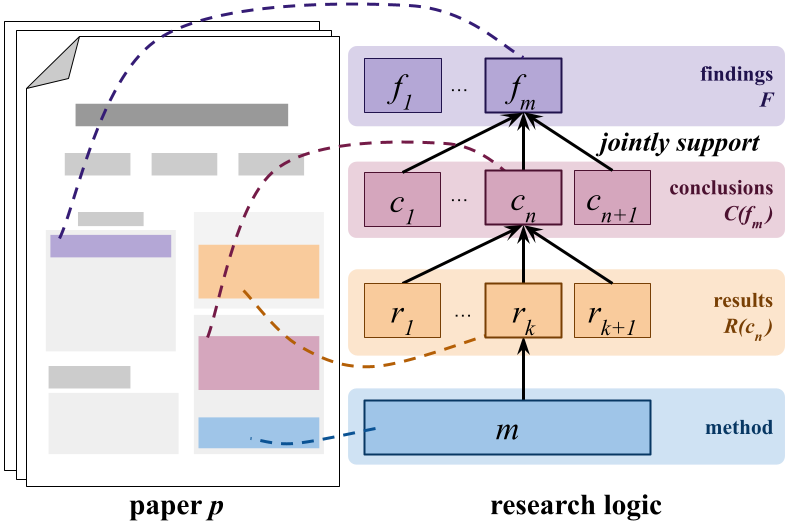}
  \caption{A paper's research logic. Each building block is evidenced in the paper (dotted lines) and jointly entail the building block above in the hierarchy (arrows).}
  \label{fig:logic}
\end{figure}

\paragraph{Soundness}
As in any empirical science, a sound NLP or AI paper makes a contribution to collective knowledge and justifies it through valid reasoning based on empirical evidence \cite{Armstrong_Green_2022}. By this definition, a paper’s soundness comprises four \textbf{building blocks}, which we refer to throughout the rest of the paper: the \textbf{method} describing the experimental setup, the empirical \textbf{results} generated by them, the \textbf{conclusions} drawn from these results, and the \textbf{finding} summarizing the conclusions as a scientific contribution. Well-designed papers explicitly present all four components. While not necessarily all papers present empirical findings, e.g., those with theoretical proofs or opinions, our analysis in \Cref{sec:dataset} shows that a vast majority of NLP and AI research papers make at least some empirical contribution. Based on this, we define our model of soundness formally. Our model effectively operationalizes \citeauthor{bacon1878novum}'s (\citeyear{bacon1878novum}) theory of the scientific method as \textit{inductive} reasoning from evidence. 

\paragraph{Formal Model}
A paper $\pap$ claims to contribute findings $\fin$. The paper $\pap$ is sound only if all findings $f \in \fin$ are sound. A finding $f$ is sound iff all underlying conclusions $\con(f)$ are sound and jointly support $f$. A conclusion $c \in \con(f)$ is sound iff its underlying results $\res(c)$ are sound and jointly support $c$. A result is sound iff the experimental methodology $\met$ adheres to best scientific practice and plausibly produces the result. We assume that $\con(f)$ and $\res(c)$ are the \textit{minimal sufficient sets} of conclusions and results, respectively; i.e., all elements are necessary to support the corresponding finding $f$ or conclusion $c$.
These building blocks form a hierarchical, logical structure, which we call the paper’s \textit{research logic}, in which arcs represent the \textit{support} relation. Figure~\ref{fig:logic} illustrates this structure. Importantly, the research logic differs from the paper’s broader argumentative structure, which consists of rhetorical moves to make the paper appear interesting and plausible \cite{teufel-etal-2009-towards}. In Aristotelian terms, the research logic refers to the logical appeal (logos) while the broader argumentation in the paper often also invokes ethos (credibility) and pathos (appeal to emotion). In other words, soundness and convincingness are independent concepts of a paper.

\paragraph{Soundness Verification}
An unsound paper violates one or more of the support relations defined above. A common case is \textit{over-claiming}, where findings misalign with the actual results; i.e., the conclusions do not jointly support the finding, usually additional conclusions would be needed. Verifying a paper’s soundness thus involves checking each building block and the corresponding support relations within the research logic hierarchy. This process relies on background knowledge only when assessing the methodology; evaluating the soundness of \textit{conclusions and findings} depends primarily on information within the paper. As such, verifying the support relationships between results, conclusions, and findings is a self-contained, context-independent task that tests reasoning and reading comprehension skills central to peer review and well-suited for evaluating ARGs.

\section{Framework} \label{sec:framework}

We use ARG as an umbrella term for systems that receive a research paper as an input and output a review report. While ARGs usually involve LLMs, we do not probe LLMs per-se but focus on their behavior in the context of automatic reviewing. A reliable ARG should detect flaws in research logic and reflect them in its output. Unlike prior work \cite[][i.a.]{zhang2025reviewing}, which checks for mentions of specific issues, we focus on the \textit{average} effect of flawed logic on automated reviews, recognizing that peer reviews weigh multiple factors and cannot highlight every concern. For instance, limited novelty might overshadow methodological flaws. 
Formally, we estimate the concept average treatment effect (ATE) \cite{goyal2019explaining} of research logic interventions on automatically generated reviews using approximate counterfactuals \cite{gat2024counterfactuals}. In other words, we surgically edit high-quality papers to compromise their research logic and quantify the resulting changes in automatic reviews, which should, on average, become more critical and emphasize soundness more compared to the automatic review of the original paper.

\subsection{Pipeline}
\begin{figure*}[t] 
  \centering
  \includegraphics[width=0.98\linewidth]{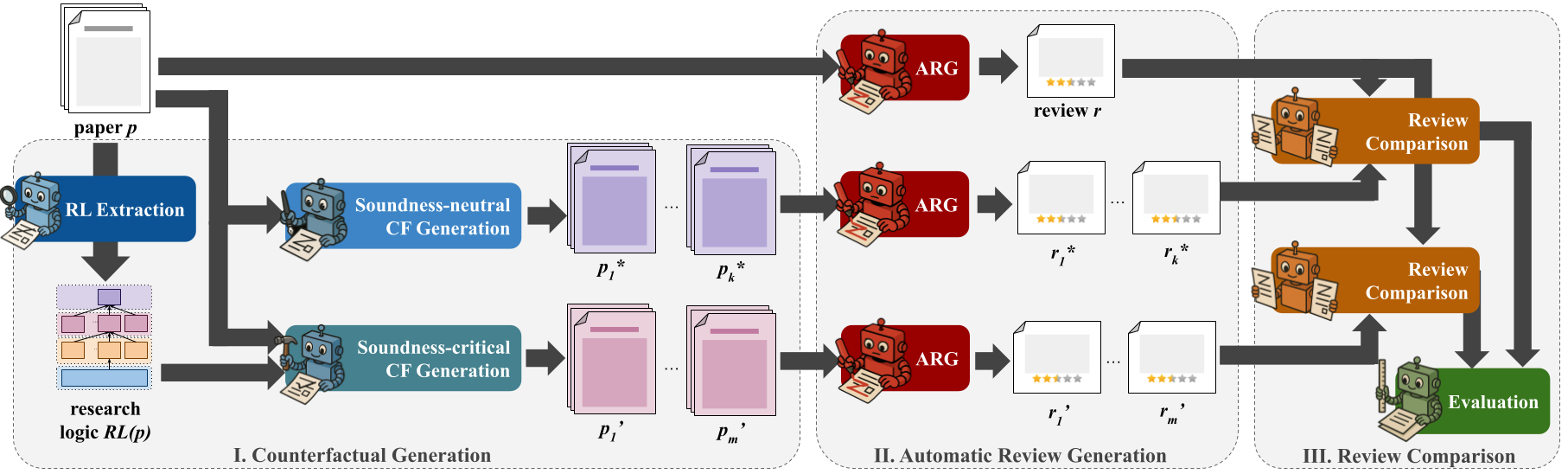}
  \caption{The counterfactual evaluation pipeline takes the original paper as an input and results in the evaluator's output. RL = research logic; CF = counterfactual; ARG = automatic review generator.}
  \label{fig:pipeline}
\end{figure*}

Our framework consists of three stages (Figure~\ref{fig:pipeline}). First, given a paper $\pap$, we generate two sets of counterfactuals (CFs): \textbf{soundness-critical} $\cfrl(p) = \{\pap'_1, ..., \pap'_m\}$, which introduces errors to the paper's research logic, altering its soundness while preserving other concepts such as clarity and novelty; and \textbf{soundness-neutral} $\cfsn =  \{p_1^*, ..., p_k^*\}$, which apply surface-level edits (e.g., formatting and language changes) to later contextualize the effects of the soundness-critical edits.

In the second stage, we run the ARG to generate a review $r$ per original paper, $\revrl = \{r_1', ..., r_m'\}$ for the soundness-critical counterfactuals, and $\revsn = \{r_1^*, ..., r_k^*\}$ for the neutral ones. In the final stage, we extract numerical features from each review and compute the differences $\deltarl(\pap)$ and $\deltasn(\pap)$ between the original review $r$ and those from $\revrl$ and $\revsn$, respectively. Aggregating over the dataset $\mathbb{D}$, we estimate the average treatment effect of soundness critical edits $\aterl(\mathbb{D}) = \frac{1}{|\mathbb{D}|} * \sum_{\pap \in \mathbb{D}}{\deltarl(p)}$, soundness-neutral edits $\atesn(\mathbb{D}) = \frac{1}{|\mathbb{D}|} * \sum_{\pap \in \mathbb{D}}{\deltasn(p)}
$, and compare them for each ARG. 

\subsection{Counterfactual Paper Generation} \label{ssec:cfgen}
Edits for counterfactual generation must satisfy four desiderata \cite{molnar2025, gat2024counterfactuals}. Edits must be \textbf{relevant (A)}; i.e., they should directly impact the paper's soundness negatively. They must be \textbf{minimal (B)}; i.e., modifying only elements tied to soundness while preserving all other concepts. Edits must also be \textbf{plausible (C)}; i.e., maintaining topical focus, fluency, and coherence. Finally, counterfactuals should be \textbf{diverse (D)}, encompassing varied modifications to the research logic.
While LLMs perform well in generating counterfactuals for sentences \cite{li2025llmcounter}, they are not readily applicable to full research papers. To address this, we propose a two-step approach (Figure \ref{fig:pipeline}, Step I). First, we automatically extract the paper's research logic. Then, LLMs disrupt the support relationship between findings, results, and conclusions (A). These changes are then projected on the paper text to ensure minimal edits (B).

\paragraph{Research Logic Extraction}
We use zero-shot prompting with one self-refinement step \cite{madaan2023self}. For each building block, we prompt the LLM to extract supporting spans with paragraph IDs and generate a summary. We apply this paradigm with building block specific adjustments for all steps.\footnote{See Appendix \ref{assec:rle} for prompts and pseudo-code.} The LLM first summarizes the paper’s research goal. It then extracts all contribution claims, filters for empirical findings, and ranks them by relevance to the goal. This ranking may not fully match human judgment but helps prioritize elements to modify during counterfactual generation. Next, it extracts all conclusions and links them to the supported findings. It identifies all results covered in figures, tables, and textual mentions and connects them to the conclusions. The output consists of the finding claims $\fin$, conclusions $\con(f)$, and results $\res(c)$, with their spans, coreferences, and joint support relations.

\paragraph{Soundness-critical Edits}
After prompting the LLM to revise a selected building block to compromise its support relations in the research logic, the LLM edits the paper reflecting the new unsound logic.
We apply three types of edits, on different levels of the research logic hierarchy for diversity (D). Table~\ref{tab:example_perturb} provides examples. For \textbf{finding edits}, we misalign a finding with its underlying conclusions drawing on prior work on misrepresented science \cite{wuehrl-etal-2024-understanding}. The LLM classifies the finding as a (i) correlational, (ii) causal, or (iii) conditional claim. If none apply, no counterfactual is generated. For correlational claims, the LLM rephrases them as causal claims; for causal claims, it inverts the causal direction; for conditional claims, it removes the stated conditions. For \textbf{conclusion edits}, we misalign a conclusion with the underlying results. The LLM first generates a hypothetical result consistent with the paper’s scope but unsupported by the experiments. It then augments the conclusion and derived finding to include this result, leaving part of it unsubstantiated. For \textbf{result edits}, we misalign a result with a derived conclusion. The LLM negates or weakens the result leading to unsupported conclusions. For example, it may reduce a strong performance gain to a marginal improvement.

We design dedicated prompts per edit type using the LLM in zero-shot mode with one self-refinement cycle. We apply the perturbations to the building blocks connected to the most important finding according to the extracted research logic to ensure high impact on the overall paper. Each operation runs independently, yielding three soundness-critical counterfactuals per paper, each creating a distinct type of issue. 
After modifying the research logic, we apply a set of zero-shot prompts with self-refinement to propagate the edits to the paper text, ensuring fluency and plausibility (C). The LLM considers the revised research logic and the associated textual spans to propose edits. As we focus on unimodal textual ARGs, edits happen only on textual mentions and captions of figures, not the figures themselves. For tables, the LLM uses a chain-of-thought prompt to reason about its structure and content before modification \cite{fang2024large}. Finally, an LLM-as-a-judge \cite{zheng2023judging} evaluates whether each counterfactual meets the desiderata. If rejected, the LLM generates a new counterfactual targeting the next top-ranked finding.

\begin{table*}[t]
    \centering
    \renewcommand{\arraystretch}{1.2}
    \footnotesize
    \begin{tabular}{P{1.8cm}P{3.8cm}P{4.4cm}P{4.2cm}}
         \toprule
          \textbf{Target}& \textbf{Original} & \textbf{Compromised} & \textbf{Paper Edit}\\
         \midrule
         \textbf{Finding} (example on \cite{lin-etal-2024-advancing}) & Spoken-LLM outperforms text-only baselines and prior speech LLM methods [...]. & Spoken-LLM outperforms all existing models [...]. & "With the same backbone model, the proposed method outperforms \textcolor{greencolor}{all existing models} [...]." \\ %
         \textbf{Conclusion} (example on \cite{chen-etal-2024-masked}) & The MFT method achieved a 5\% increase in accuracy on the GSM8K dataset. & The MFT method achieved a 5\% increase in accuracy on the GSM8K dataset, with an even greater improvement of 7\% observed [...]. &  "With just this minor modification, a 5\% increase in accuracy can be achieved [...] \textcolor{greencolor}{and an even greater improvement of 7\% [...]}." \\ %
         
         \textbf{Result} (example on \cite{rao-etal-2023-chatgpt}) & The consistency scores [...] were quantified, revealing that ChatGPT had a score of 0.907 [...]. & Our findings indicate that while ChatGPT's consistency score was slightly lower at 0.807 compared [...] & Table 2: \textcolor{redcolor}{0.907} $\rightarrow$ \textcolor{greencolor}{0.807}\\ %
         \bottomrule
    \end{tabular}
    \caption{Examples of perturbations targeting different building blocks from our dataset (shortened for brevity). We show the original building block, the compromised version, and resulting paper edit.}
    \label{tab:example_perturb}
\end{table*}

\paragraph{Soundness-neutral Edits}
We apply four simple edits that preserve both content and soundness. We randomly select a subset of paragraphs for each edit type. For \textbf{active-to-passive}, the LLM rewrites text from active to passive voice and vice versa. For \textbf{American-to-British}, it converts spelling between American and British English. For \textbf{language error}, we inject minor spelling and grammar mistakes. For \textbf{paper layout}, we multiply whitespaces at random and relocate figure captions and tables to the end of the paper, altering layout but not content.

\subsection{Review Comparison} \label{ssec:delta}
The goal of third step (see Figure~\ref{fig:pipeline}) is to quantify differences between reviews of the original and counterfactual papers. Faulty research logic should lead to reviews that are, on average, more negative and more focused on soundness. We approximate these concepts using three review features computing the difference per feature between the counterfactual $r_c$ and the original review $r_o$ as the \textbf{review difference}.
First, we analyze the distribution of \textbf{aspects}, i.e. the paper dimensions  that the review discusses, such as experiments or presentation. We detect aspects using a RoBERTa model fine-tuned by \citet{lu2025identifying}, and compute the number of research-logic-related aspects. We compute the review difference as $\textit{aspect}(r_c, r_o)=\textit{\#aspects}(r_c) - \textit{\#aspects}(r_o)$. A large positive value indicates an increased focus on soundness as this means that soundness-related comments were added to the review.
Second, we assess the \textbf{sentiment} of review assertions. Reviews consist of positive, negative, or neutral assertions on the paper \cite{dycke2023overview}. We extract and classify these assertions by their sentiment using GPT-4o-mini in zero-shot mode with self-refinement. We compute the review difference as $\textit{sentiment}(r_c, r_o)=\%\textit{positive assertions}(r_c) - \%\textit{positive assertions}(r_o)$ in $[-1,1]$. Values closer to $-1$ mean stronger criticism related to soundness.
Third, aligning with \citet{li2025aspect}, we track changes in the \textbf{review score} summarizing the overall reviewer opinion. We compute $\textit{score}(r_c, r_o) = \textit{score}(r_c) - \textit{score}(r_o)$. We expect the score to drop when soundness is compromised as indicated by a negative value.

\section{Dataset} \label{sec:dataset}
Applying our pipeline, we first create a dataset of papers, associated research logic, and counterfactual versions, and validate each step. We list the model versions in Appendix \ref{assec:cfgen} and prompts in the supplementary materials.
We develop prompts for each step individually and manually verify outputs on a subset of papers. We begin with a basic prompt and iteratively refine it using Claude Sonnet 3.5 \cite{anthropic2024claude35sonnet} akin to meta-prompting \cite{zhou2023large}. We adjust prompts until we consider the outputs correct for all test samples. All prompts follow a similar structure and use \texttt{json} output format in line with prompt engineering best practices \cite{phoenix2024prompt} (see \autoref{fig:general_prompt} in the Appendix).

\subsection{Source Data}
We construct our dataset from four major AI and NLP conferences over two years to generalize beyond individual venues \cite{kuznetsov2024can}. Specifically, we include papers from the Association of Computational Linguistics (ACL) conferences ACL-23 and ACL-24, EMNLP-23 and EMNLP-24, from the Neural Information Processing Systems conference NeurIPS-24, and International Conference on Learning Representations ICLR-25.
To ensure high-quality input papers with valid research logic, we include only accepted papers. We collect openly licensed papers from ICLR-25, NeurIPS-24, and EMNLP-23 via OpenReview\footnote{\url{https://openreview.net}}, and from the ACL Anthology\footnote{\url{https://aclanthology.org/}} for the remaining ACL conferences, yielding approximately $13k$ instances.
We convert papers to Markdown format suitable for LLM processing. We remove instances with parsing issues retaining $3,532$ reliably parsed papers. From these, we sample $156$ papers evenly across conferences: $18$ for development and prompt tuning, and $138$ for testing. This sample size balances paper diversity with the computational cost of review generation for original and counterfactual papers later. Appendix \ref{asec:preproc} details the preprocessing procedure.

\subsection{Research Logic Data}
Through manual refinement, we develop the prompts for extracting the research logic on the development set and select the best performing model. GPT-4o-mini \cite{openai2024gpt4technicalreport} offered the best trade-off between inference speed, cost and output quality. We employ self-refinement \cite{madaan2023self}, which notably improved the alignment between building blocks consistent with findings on other reasoning-intensive tasks \cite{sahoo2024systematic}. A single refinement cycle achieved the best performance–cost trade-off. We query the OpenAI API\footnote{\url{https://platform.openai.com/}} for model inference. On average, the extraction of the research logic takes roughly five minutes per paper.

\paragraph{Validation Study}
We validate the accuracy of research logic extraction through human evaluation focusing on the factual alignment between each extracted building block and the source paper.
We recruit three postgraduate NLP researchers with at least four years of experience reading academic papers. For each annotation, we provide the paper, the building block type, its summary, and the main passage provided by the LLM. Annotators assess the factual correctness of the summary, focusing on the main highlighted passage but they are allowed to consult the full paper. We also ask them to note issues beyond factual accuracy.
We pilot the process on $50$ randomly sampled building blocks from $5$ papers. We then compute inter-annotator agreement (IAA) on this pilot plus $10$ additional samples. In total, annotators label $139$ building blocks from $14$ papers across conferences.
We use Goldstein’s S \cite{bennett1954communications} to measure IAA, as it is more suited for imbalanced label distributions than Krippendorff’s $\alpha$ \cite{feinstein1990high}, given that $74\%$ of labels are positive. We obtain $S=0.37$, indicating moderate agreement. Upon inspection, we find that in $12\%$ of all samples the LLM selects an incorrect passage to support its summary; in $56\%$ of disagreement cases, at least one annotator flags this issue. This induces disagreement because of varying considered context by the annotators.

\paragraph{Validation Result}
We compute the majority vote for redundant annotations and merge them with individually labeled instances, finding that $77\%$ of building blocks are factually accurate. We consider this accuracy sufficient as an intermediate step in counterfactual generation. To address the issue of incorrect evidence spans, we revise the extraction prompts requiring the LLM to cite specific text spans together with their paragraphs ID and apply it throughout the remainder of the study.

\subsection{Counterfactual Data}
Following our general prompt development paradigm, we tune prompts and select the best LLM based on manual inspection on the development set.

\paragraph{Soundness-neutral Counterfactuals}
We generate soundness-neutral counterfactuals using Phi-4 14B \cite{abdin2024phi}. We run Phi-4 (with \texttt{Q4\_K\_M} quantization) on two L40 GPUs with approximately 14GB effective memory use during inference. On average, the generation of one soundness-neutral counterfactual takes roughly three minutes. For active-to-passive and American-to-British, we randomly select 40\% of the paragraphs; for language error, we select 20\% to simulate minor language issues. Given the simplicity of these edits, one author manually inspects $10$ counterfactuals per type to verify that the edits do not affect the paper content. In all cases, the revised text preserves the original meaning. 

\paragraph{Soundness-critical Counterfactuals}
We use GPT-4o-mini to generate soundness-critical counterfactuals. During prompt development, we found that decomposing the process into multiple stages, where each corresponds to a single LLM call producing an intermediate result with an explanation, substantially improved output quality. We explicitly separate stages that require reasoning from those that modify the paper’s content since this produced more diverse outputs. We further divide creative stages, such as proposing new hypothetical results (see Sec. \ref{ssec:cfgen}), into candidate generation and subsequent selection.
If an edit type is not applicable, e.g., a paper lacks the necessary claim types for finding edits, we skip the counterfactual for that paper. 
The generation of one soundness-critical counterfactual takes on average roughly two minutes using the OpenAI API.

\paragraph{Validation Study}
The human validation assesses whether counterfactuals meet the desiderata (see \Cref{ssec:cfgen}). We employ the same three postgraduate NLP researchers as in the previous validation.
We provide annotators with the counterfactual paper, the extracted research logic, the edited building block, and highlight the edits in the paper. First, annotators judge if the research logic is compromised (RL-cor) and plausible within the paper's scope (RL-pla). Then, they evaluate the edits applied to the paper considering whether the edits affect the soundness (E-cor), are plausible (E-pla), and minimal (E-min).
We pilot the annotation on $7$ counterfactuals. For the main study, annotators assess $92$ counterfactuals from $10$ papers, including $25$ used to compute IAA. Due to label imbalance, we report Goldstein’s S: $0.46$ for RL-cor, $0.57$ for RL-pla, $0.20$ for E-cor, $0.62$ for E-pla, and $0.57$ for E-min.
Annotators show moderate to substantial agreement on the research logic modifications. In contrast, assessing the correctness of paper edits proves more subjective, particularly for E-cor; this is consistent to levels of subjectivity in peer review \cite{bornmann_scientific_2011}. Disagreements occur more frequently for edits of conclusions ($40\%$) and findings ($33\%$) where annotator comments indicate that certain edits, such as inserting the word 'significant' as a statistical claim, are ambiguous leading to disagreement.

\paragraph{Validation Result}
We use the majority vote on redundant annotations merged with individually labelled instances. Annotators judge $88\%$ of research logic edits as correct (RL-cor) and $91.3\%$ as plausible (RL-pla). They find that $90.2\%$ of the edits compromise the paper’s soundness (E-cor), $91.3\%$ are plausible (E-pla), and $79.4\%$ are minimal (E-min). For the purpose of our evaluation, we consider the desiderata met and account for residual noise during evaluation by comparing soundness-neutral and soundness-critical counterfactuals on a large set of instances.

\subsection{Dataset Overview}

\begin{table}[t]
\centering
\small
\begin{tabular}{r c}
\toprule
\multicolumn{2}{c}{\textbf{Paper Distribution}} \\
\midrule
\textbf{\#papers} & $133$ \\
\textbf{\#papers p. conference} & $22.50 \pm 1.80$ \\
\textbf{\#papers p. institution} & $1.38 \pm .08$\\
\midrule
\multicolumn{2}{c}{\textbf{Research Logic Distribution}} \\
\midrule
\textbf{\#paper types} & $7$ \\
\textbf{\#papers p. paper type} & $19.29 \pm 33.68$ \\
\textbf{\#findings p. paper} & $3.69 \pm 1.93$ \\
\bottomrule
\end{tabular}
\caption{Dataset statistics for the papers and their extracted research logic in the test set.}
\label{tab:underlying_dataset}
\end{table}

\begin{table}[t] %
\centering
\small
\begin{tabular}{r c c c}
\toprule
& \textbf{CF}$_{\bm{CR}}$ & \textbf{CF}$_{\bm{NE}}$ & \textbf{total}\\
\midrule
\textbf{\#CFs} & $391$ & $540$ & $931$ \\
\textbf{\#CFs / paper} & $2.9$ & $4.0$ & $7.0$\\
\midrule
\textbf{\#edits / CF} & $2.59_{\pm 1.59}$ & $5.81_{\pm 1.65}$ & $4.23_{\pm2.28}$ \\
\textbf{diff. / CF} & $579_{\pm 1159}$ & $771_{\pm 1003}$ & $677_{\pm1087}$\\
\bottomrule
\end{tabular}
\caption{Dataset statistics for soundness-critical ($\cfrl$) and soundness-neutral ($\cfsn$) counterfactuals. The number of edits counts edit operations; diff. is the Levensthein distance.}
\label{tab:dataset}
\end{table}

\autoref{tab:underlying_dataset} and \ref{tab:dataset} summarize the final evaluation dataset of $931$ counterfactuals on $133$ papers, excluding $5$ papers without empirical findings. 

\paragraph{Diversity of Counterfactuals}
As shown in \autoref{tab:dataset}, the number of edits and amount of changed text is comparable for both counterfactual types, ensuring fair comparison.
Among soundness-critical edits, $85.6\%$ of them appear in the text, $9.6\%$ in tables, and $4.7\%$ in figure captions; the latter two only occur for result edits. The dominance of text edits reflects the textual nature of findings and conclusions, and the fact that many results are reported and interpreted in the text. Each edit type engages distinct reasoning over different areas of the paper.

\paragraph{Diversity of Papers}
As shown in \autoref{tab:underlying_dataset}, the papers are nearly evenly distributed across conferences. To assess authorship diversity, we use the last author’s affiliated institution as a proxy, retrieving metadata via the OpenAlex API\footnote{\url{https://openalex.org/}} (excluding $10$ missed authors). On average, $1.38$ papers originate from the same institution, indicating high diversity. Regarding research logic, of the $7$ paper types classified during extraction, papers introducing new methodology dominate ($75\%$), followed by analysis ($12\%$) and dataset papers ($6\%$). This skew towards methodological work likely reflects the actual distribution of publications in AI and NLP conferences and encourages future work on more diverse paper types (see Sec. \ref{sec:limit}).

\section{Evaluation}
We run the evaluation pipeline based on the previously generated dataset to estimate the average treatment effect of soundness-neutral and -critical edits and compare them. The detailed model versions, generation hyperparameters, and prompts are reported in Appendix \ref{asec:experiments}.

\subsection{Experiments}
\paragraph{ARGs}
We consider three types of ARGs from the literature.
The first type uses LLMs in zero-shot mode either with a generic review prompt (\textsc{Zero-Generic}), e.g. in \citet{liang2024can}, or a prompt including venue-specific guidelines (\textsc{Zero-Guide}), e.g. in \cite{du-etal-2024-llms}. We test both with GPT-4o-mini and GPT-4.1 as large proprietary LLMs, DeepSeek-14B and DeepSeekV3 \cite{liu2024deepseek} as reasoning LLMs, and Phi-4 as a small open-weight model. We design the prompts based on related literature to ensure alignment, while adding a standardized output format for easy parsing. We perform formatting and plausibility checks on the development set without further tuning.
The second type comprises multi-agent systems where LLMs specialize in different paper aspects and engage in a discussion. We include TreeReviewer \cite{chang2025treereview}, implemented with GPT-4o-mini, as a representative. MARG \cite{d2024marg}, evaluated on our development set, proved computationally infeasible with individual reviews requiring at least $20$ minutes and up to $2$ hours.
The third type uses LLMs fine-tuned on peer review data. We test Reviewer2 \cite{gao2024reviewer2}, fine-tuned with an automatic prompting model, and DeepReviewer \cite{zhu2025deepreview}, which is further trained on synthetic reasoning data.
We use default hyperparameters for all ARGs. We fix the random seed and, for zero-shot LLMs, set the temperature to zero to minimize output variance. We truncate the paper to the effective context window size per LLM \cite{hsieh2024ruler}. When ARGs produce semi-structured text, e.g. DeepReviewer, we parse the output using regular expressions and, if needed, fall back to GPT-4o-mini for parsing. We include reviews that cannot be parsed into the venue's template in raw form; if the parsing of scores fails, we do not consider those scores for evaluation.

\paragraph{Oracle Ablation}
To validate, we include an \textsc{oracle} ARG. \textsc{oracle} takes the reviews generated by \textsc{Zero-Guide-GPT-4om} for the original papers and paraphrases them for each counterfactual. For soundness-critical counterfactuals, it also adds a comment on the introduced issue and lowers the overall score randomly. This setup simulates an ARG that reports the soundness issue in the review, and adjusts the overall score as a mild but notable reaction to unsound research logic.

\paragraph{Statistical Analysis}
For each review difference dimension, we fit a linear mixed effect model (LME) \cite{lindstrom1988newton} per ARG. We test the null hypothesis that the ATE of the two conditions, soundness-critical or soundness-neutral, is identical; in other words, we test if soundness-critical edits have no stronger effect on reviews than the soundness-neutral edits. LMEs are designed for varying repeated measures. In our case, there are multiple review differences per paper; we model the paper as a random effect, while the review difference and condition are fixed effects. Finally, we use Benjamin-Hochberg correction \cite{benjamini1995controlling} due to multiple testing on ARGs. 

\subsection{Results}

\begin{figure}[t] 
  \centering
  \includegraphics[width=\linewidth]{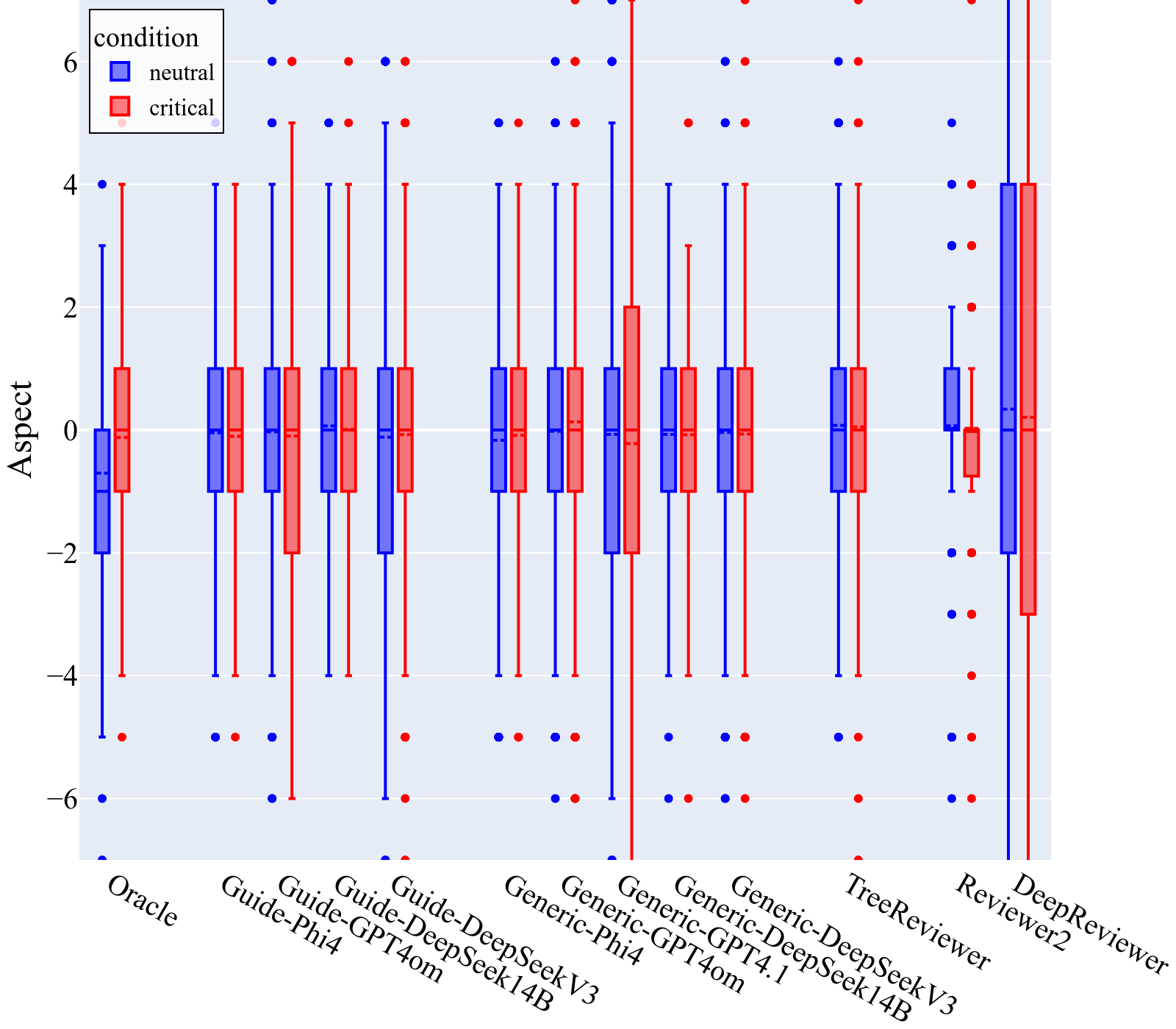}
  \caption{ATE of \textbf{aspect differences} for soundness-critical and -neutral CFs; the larger the difference of neutral/critical means (dashed horizontal lines) the better.}
\label{fig:result1}
\end{figure}

\begin{figure}[t] 
  \centering
  \includegraphics[width=\linewidth]{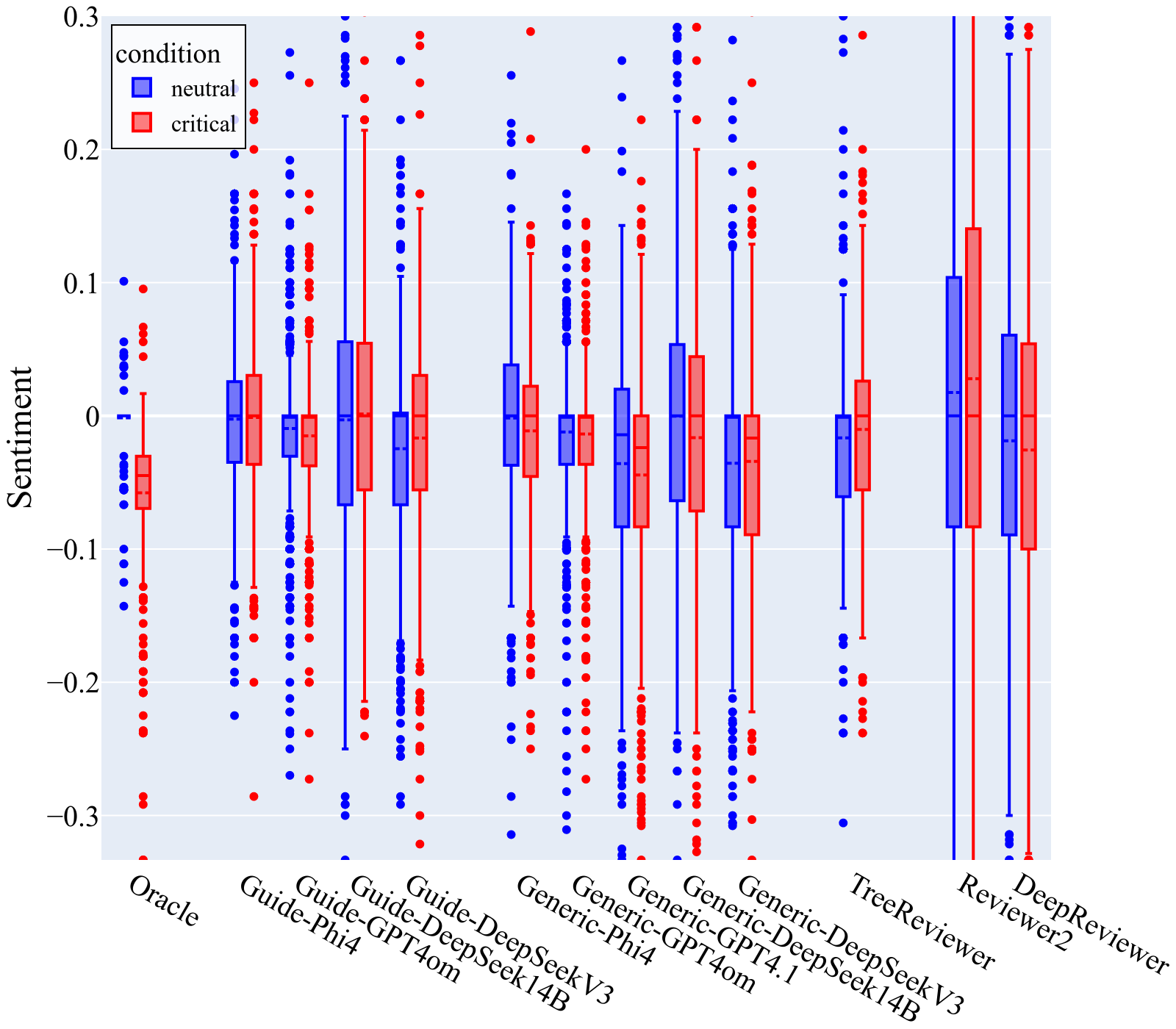}
  \caption{ATE of \textbf{sentiment differences} for soundness-critical and -neutral CFs; the larger the difference of neutral/critical means (dashed horizontal lines) the better.}
\label{fig:result2}
\end{figure}

\begin{figure}[t] 
  \centering
  \includegraphics[width=\linewidth]{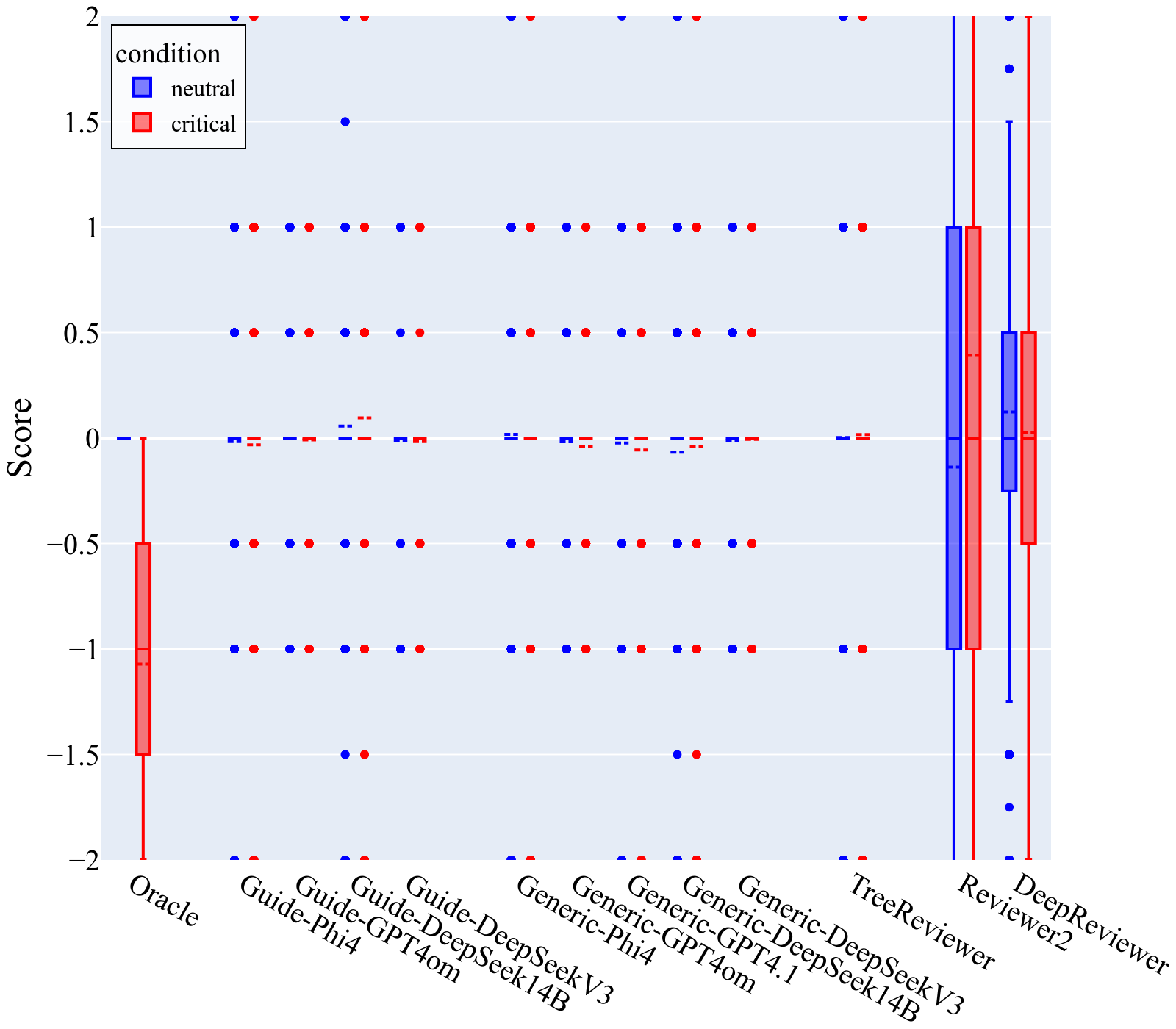}
  \caption{ATE of \textbf{score differences} for soundness-critical and -neutral CFs; the larger the difference of neutral/critical means (dashed horizontal lines) the better.}
\label{fig:result3}
\end{figure}

\paragraph{Are automatic review reports significantly affected by soundness-critical edits?}

\autoref{fig:result1}, \ref{fig:result2}, and  \ref{fig:result3} show the box plots of the ATE per difference dimension and both counterfactual types per ARG. To recap, 'Aspect' (\autoref{fig:result1}) measures the number of soundness-related statements, 'Sentiment' (\autoref{fig:result2}) the density of positive comments, and 'Score' (\autoref{fig:result3}) the  overall rating; all of them compare the ARG's reviews for the counterfactuals with the one for the original (see \Cref{ssec:delta}).
For an ideal ARG, the ATE for soundness-critical and -neutral edits should lie far apart on all dimensions; the ATE of critical edits should be positive for aspects (soundness-related aspects are added) and negative for sentiment (less positive comments) and score (lower rating).

As expected, the ATE of soundness-neutral edits are close to zero for most ARGs. The variance is large for all ARGs and especially for the fine-tuned Reviewer2 and DeepReviewer. However, for all ARGs the ATE of soundness-critical edits is similar to the ATE of neutral edits. At a significance level of $\alpha = 0.05$, none of the models show a significant difference in the ATE between neutral and critical edits for aspects, sentiment, or score.\footnote{Detailed p-values are reported in Appendix \Cref{tab:pvalues}.} \textbf{In other words, faulty research logic has no statistically significant impact on the automatically generated reviews of any tested ARG.}

\paragraph{Which of the ARGs are influenced most by faulty research logic?} 
Although none of the ARGs are significantly affected by flawed research logic, we rank the models to identify tendencies. A coherent ARG should show low ATE variance for soundness-neutral counterfactuals, being resilient to surface-level changes, and a large difference between the ATE for neutral and critical counterfactuals. To capture this in a single score, we express the ATE difference between both conditions as a multiple of the effect standard deviation of the neutral counterfactuals. For each ARG, we compute $z = \frac{|\aterl - \atesn|}{\sigma(\atesn)}$ per dimension and then average over all dimensions. This yields the ranking shown in \Cref{tab:ranking}. Notably, generic zero-shot prompts perform better than those with guidance, which may act as a distractor. For Phi-4, with its smaller context window, \textsc{Zero-Guide} likely causes truncation of key paper content. While the fine-tuned Reviewer2 ranks highest, DeepReviewer ranks lowest; likely, their backbone models, Llama-2 for Reviewer2 and Phi-4 for DeepReviewer, play an important role since the \textsc{Zero-Guide} results for Phi-4 suggest issues with context size and distractors. This warrants further investigation in future work. Finally, there is no clear link between model size and performance, as Phi-4 and GPT-4.1 both rank highly.

\paragraph{How does an ideal ARG perform?}
To validate the evaluation pipeline, we evaluate the \textsc{oracle} ARG along with the others. Here, for all three dimensions there is a notable difference between critical and neutral counterfactuals which is confirmed by statistical analysis. The \textsc{oracle} reviews are significantly influenced (at $\alpha = 0.05$) by faulty research logic along all dimensions hereby confirming the validity of our pipeline.

\paragraph{How do these results contextualize with prior work?} 
Our results suggest that prior reports of ARGs achieving error discovery rates of around $40\%$ for GPT-4o-mini \cite{zhang2025reviewing} are influenced by the types of errors tested, which often rely on background knowledge recall; a task at which LLMs perform well \cite{alkhamissi2022review}. For example, spotting an error in a softmax formula mainly tests recall with minimal reasoning. In contrast, our experiments isolate \textit{reasoning} and show that ARGs clearly lack this skill. From this, we draw two recommendations for future work. First, to accurately evaluate ARGs, \textbf{studies should test distinct skills in isolation to identify specific limitations and capabilities}. Second, our findings \textbf{highlight the potential for human–LLM collaboration in peer review.}
Our results show that ARGs alone fail to assess the consistency of research logic, whereas human reviewers benefit from AI assistance in knowledge-intensive steps \cite{dycke2025strictastructuredreasoningcritical}. This suggests that peer review is well-suited to human–AI collaboration through co-construction \cite{dutta2025problemsolvinghumanaipreferencebased}, where a human expert and an AI system jointly construct a review through iterative interaction, compensating for each other’s limitations.

\begin{table}[t] %
\centering
\footnotesize
\begin{tabular}{r c}
\toprule
\textbf{ARG} & \textbf{z} \\
\midrule
\textsc{oracle} & $1.702$ \\
\midrule
Reviewer2 & $0.126$ \\
\textsc{Zero-Generic-GPT4.1} & $0.078$ \\
\textsc{Zero-Generic-Phi4} & $0.072$ \\
\textsc{Zero-Generic-GPT4om} & $0.063$ \\
\textsc{Zero-Generic-DeepSeek14B} & $0.063$ \\
\textsc{Zero-Guide-GPT4om} & $0.053$ \\
\textsc{Zero-Guide-DeepSeek14B} & $0.051$ \\
\textsc{Zero-Guide-DeepSeekV3} & $0.044$ \\
DeepReviewer & $0.042$ \\
\textsc{TreeReviewer} & $0.040$ \\
\textsc{Zero-Guide-Phi4} & $0.029$ \\
\textsc{Zero-Generic-DeepSeekV3} & $0.018$ \\
\bottomrule
\end{tabular}
\caption{Ranking of the ARGs by score $z$.}
\label{tab:ranking}
\end{table}

\paragraph{How sensitive are ARGs to the paper's surface form?}  
Notably, all ARGs show high standard deviations for soundness-neutral counterfactuals, suggesting that ARGs are sensitive to spurious features similar to issues of prompt sensitivity in LLMs \cite{sclar2024quantifying}. To examine this further, we measure lexical similarity using ROUGE-2 and use GPT-4o-mini to detect whether assertions align between the original and soundness-neutral reviews, computing the Jaccard index on assertions. We perform human validation of the automatic alignment with two annotators on $50$ assertion pairs with substantial agreement (Krippendorff's $\alpha = 0.74$) attesting an accuracy of $0.78$. We analyze $48$ randomly sampled papers (roughly 8 per conference). Table \ref{tab:surface} in the Appendix reports the detailed results.
Surface similarity ranges from $0.25$ ROUGE-2 for DeepReviewer to $0.54$ for \textsc{Zero-Guide-GPT4om}, indicating substantial changes in wording for many ARGs. This pattern holds for the assertion overlap, with a Jaccard index from $0.27$ for DeepReviewer to $0.57$ for \textsc{Zero-Generic-GPT4om} indicating low to moderate overlap in the contents of the reviews. Overall, GPT-4o-mini produces more similar reviews than other ARGs, regardless of the prompting strategy. Even allowing for some measurement error, ARGs remain highly sensitive to spurious surface changes in the input text.
Our findings point to another recommendation for future work: \textbf{all ARG evaluations, whether using human reference data or sensitivity analysis, should report mean performance on multiple soundness-neutral versions of the papers} akin to prompt-sensitivity reporting to better reflect their true capability independent of spurious features. Here, our counterfactual dataset may also help improve model consistency through training data augmentation \cite{wu-etal-2021-polyjuice}.

\section{Limitations and Future Work} \label{sec:limit}
In this Section, we summarize potential limitations of the proposed framework and point to important future work.

\paragraph{Assumptions}
We design the counterfactual generation to be generic and applicable across diverse papers. However, our approach makes assumptions on the paper structure that may not hold for all paper types and domains. Future work should extend faulty research logic detection to additional domains, paper types, and logical structures.
For evaluation, we use accepted papers from reputable, peer-reviewed  venues to ensure the original research logic is sound and extractable. While this assumption may not hold for every paper, averaging effects over a large, diverse sample helps manage noise. Future extensions to new data crucially need quality assurance of the underlying papers.

\paragraph{Counterfactual Generation}
For counterfactual generation, we validate the plausibility and diversity of edits aiming to resemble \textit{human} research logic errors. However, the distribution of constructed errors likely differs from those in real papers. Our evaluation tests ARGs’ canonical ability to detect flawed research logic under controlled conditions. If an ARG’s output is unaffected by flawed logic in this setting, it will also fail on real papers. Conversely, a significant effect does not guarantee strong real-world performance underscoring the need to study research logic flaws in human papers.
Finally, we focus on unimodal ARGs, as editing figures poses a separate challenge and requires distinct reviewing skills \cite{llmchartunderstanding2025}. Instead, we modify figure captions and related text to simulate figure edits from a text-only reviewer’s perspective. Future research should explore counterfactual generation for figures and test multimodal ARGs.

\paragraph{Interpretation}
Our experiments show that state-of-the-art ARGs fail to identify flawed research logic in research papers. We further find that ARGS are sensitive to surface-level features, suggesting that the underlying LLMs rely on superficial heuristics instead of thorough reasoning. Investigating the causes of this behavior is an important direction for future work. In particular, distinguishing failures to identify soundness-relevant information in the paper from failures to recognize logical fallacies within the research logic itself are crucial for developing more robust ARGs in the future.

\paragraph{Reproducibility}
For dataset creation and evaluation we primarily employ closed commercial models, as their outputs demonstrated higher quality during manual tuning. To enhance replicability, we release all model outputs in the supplementary materials. For reproducibility, detailed model versions are reported in Appendix \ref{asec:hyper}. However, the use of closed models inevitably limits future reproduction on new data, as such models may undergo unrecorded changes. In future work, the extension to open models that perform better or on-par with commercial LLMs for these steps is a promising avenue to enhance reproducibility.

\newpage

\section{Conclusion}
We introduced a novel, fully automatic counterfactual evaluation framework for ARGs, focusing on the detection of flawed research logic. We proposed a three-step pipeline to estimate the effect of soundness-critical edits to papers on automatically generated peer reviews, testing the reasoning capabilities of ARGs. Our results show that current ARGs fail to detect faulty research logic. Based on this, we propose three directions to advance LLM-based reviewing systems: designing dedicated tests for distinct reviewing skills, fostering human–machine collaboration, and accounting for sensitivity to surface-level paper features during evaluation. Our work lays the groundwork for robust evaluation of ARGs.

\section*{Acknowledgements}
This work has been funded by the LOEWE Distinguished Chair “Ubiquitous Knowledge Processing”, LOEWE initiative, Hesse, Germany (Grant Number: LOEWE/4a//519/05/00.002(0002)/81).
This work has been co-founded by the German Federal Ministry of Education and Research and the Hessian Ministry of Higher Education, Research, Science, and the Arts within their joint support of the National Research Center for Applied Cybersecurity ATHENE. 
This research work has been co-funded by the European Union (ERC, InterText, 101054961). Views and opinions expressed are, however, those of the author(s) only and do not necessarily reflect those of the European Union or the European Research Council. Neither the European Union nor the granting authority can be held responsible for them. 
We express our sincere gratitude to the reviewers and action editor at TACL for their valuable and constructive feedback.

\bibliography{tacl2021}

\begin{thebibliography}{72}
\expandafter\ifx\csname natexlab\endcsname\relax\def\natexlab#1{#1}\fi

\bibitem[{Abdin et~al.(2024)Abdin, Aneja, Behl, Bubeck, Eldan, Gunasekar, Harrison, Hewett, Javaheripi, Kauffmann, Lee, Lee, Li, Liu, Mendes, Nguyen, Price, de~Rosa, Saarikivi, Salim, Shah, Wang, Ward, Wu, Yu, Zhang, and Zhang}]{abdin2024phi}
Marah Abdin, Jyoti Aneja, Harkirat Behl, Sébastien Bubeck, Ronen Eldan, Suriya Gunasekar, Michael Harrison, Russell~J. Hewett, Mojan Javaheripi, Piero Kauffmann, James~R. Lee, Yin~Tat Lee, Yuanzhi Li, Weishung Liu, Caio C.~T. Mendes, Anh Nguyen, Eric Price, Gustavo de~Rosa, Olli Saarikivi, Adil Salim, Shital Shah, Xin Wang, Rachel Ward, Yue Wu, Dingli Yu, Cyril Zhang, and Yi~Zhang. 2024.
\newblock \href {https://arxiv.org/abs/2412.08905v1} {Phi-4 technical report}.
\newblock \emph{arXiv preprint arXiv:2412.08905v1}.

\bibitem[{AlKhamissi et~al.(2022)AlKhamissi, Li, Celikyilmaz, Diab, and Ghazvininejad}]{alkhamissi2022review}
Badr AlKhamissi, Millicent Li, Asli Celikyilmaz, Mona Diab, and Marjan Ghazvininejad. 2022.
\newblock \href {https://arxiv.org/abs/2204.06031v1} {A review on language models as knowledge bases}.
\newblock \emph{arXiv preprint arXiv:2204.06031v1}.

\bibitem[{Anthropic(2024)}]{anthropic2024claude35sonnet}
Anthropic. 2024.
\newblock Claude 3.5 sonnet.
\newblock \url{https://www.anthropic.com/news/claude-3-5-sonnet}.
\newblock Accessed: 2025-11-05.

\bibitem[{Armstrong and Green(2022)}]{Armstrong_Green_2022}
John~Scott Armstrong and Kesten~C. Green. 2022.
\newblock \emph{The Scientific Method: A Guide to Finding Useful Knowledge}.
\newblock Cambridge University Press.

\bibitem[{{Association for the Advancement of Artificial Intelligence}(2025)}]{aaai2025aiPeerReview}
{Association for the Advancement of Artificial Intelligence}. 2025.
\newblock {AAAI} launches {AI}‑powered peer review assessment system.
\newblock \url{https://aaai.org/aaai-launches-ai-powered-peer-review-assessment-system/}.
\newblock Accessed: 2025-07-18.

\bibitem[{Bacon(1878)}]{bacon1878novum}
Francis Bacon. 1878.
\newblock \emph{Novum Organum}.
\newblock Clarendon press.

\bibitem[{Benjamini and Hochberg(1995)}]{benjamini1995controlling}
Yoav Benjamini and Yosef Hochberg. 1995.
\newblock Controlling the false discovery rate: {A} practical and powerful approach to multiple testing.
\newblock \emph{Journal of the Royal statistical society: series B (Methodological)}, 57(1):289--300.

\bibitem[{Bennett et~al.(1954)Bennett, Alpert, and Goldstein}]{bennett1954communications}
Edward~M. Bennett, Renee Alpert, and A.~C. Goldstein. 1954.
\newblock Communications through limited response questioning.
\newblock \emph{Public opinion quarterly}, pages 303--308.

\bibitem[{Birukou et~al.(2011)Birukou, Wakeling, Bartolini, Casati, Marchese, Mirylenka, Osman, Ragone, Sierra, and Wassef}]{birukou2011peerreview}
Aliaksandr Birukou, Joseph~R. Wakeling, Claudio Bartolini, Fabio Casati, Maurizio Marchese, Katsiaryna Mirylenka, Nardine Osman, Azzurra Ragone, Carles Sierra, and Aalam Wassef. 2011.
\newblock \href {https://doi.org/10.3389/fncom.2011.00056} {Alternatives to peer review: Novel approaches for research evaluation}.
\newblock \emph{Frontiers in Computational Neuroscience}, volume 5 - 2011.

\bibitem[{Bornmann(2011)}]{bornmann_scientific_2011}
Lutz Bornmann. 2011.
\newblock \href {https://doi.org/10.1002/aris.2011.1440450112} {Scientific peer review}.
\newblock \emph{Annual Review of Information Science and Technology}, 45(1):197--245.

\bibitem[{Chamoun et~al.(2024)Chamoun, Schlichtkrull, and Vlachos}]{chamoun-etal-2024-automated}
Eric Chamoun, Michael Schlichtkrull, and Andreas Vlachos. 2024.
\newblock \href {https://doi.org/10.18653/v1/2024.findings-acl.580} {Automated focused feedback generation for scientific writing assistance}.
\newblock In \emph{Findings of the Association for Computational Linguistics: ACL 2024}, pages 9742--9763, Bangkok, Thailand. Association for Computational Linguistics.

\bibitem[{Chang et~al.(2025)Chang, Li, Zhang, Kong, Wu, So, Guo, Zhu, and Wong}]{chang2025treereview}
Yuan Chang, Ziyue Li, Hengyuan Zhang, Yuanbo Kong, Yanru Wu, Hayden Kwok-Hay So, Zhijiang Guo, Liya Zhu, and Ngai Wong. 2025.
\newblock \href {https://doi.org/10.18653/v1/2025.emnlp-main.790} {{T}ree{R}eview: A dynamic tree of questions framework for deep and efficient {LLM}-based scientific peer review}.
\newblock In \emph{Proceedings of the 2025 Conference on Empirical Methods in Natural Language Processing}, pages 15662--15693, Suzhou, China. Association for Computational Linguistics.

\bibitem[{Chen et~al.(2024)Chen, Wang, Lin, Lv, Wu, Gao, Wen, Yan, and Li}]{chen-etal-2024-masked}
Changyu Chen, Xiting Wang, Ting-En Lin, Ang Lv, Yuchuan Wu, Xin Gao, Ji-Rong Wen, Rui Yan, and Yongbin Li. 2024.
\newblock \href {https://doi.org/10.18653/v1/2024.acl-long.320} {Masked thought: Simply masking partial reasoning steps can improve mathematical reasoning learning of language models}.
\newblock In \emph{Proceedings of the 62nd Annual Meeting of the Association for Computational Linguistics (Volume 1: Long Papers)}, pages 5872--5900, Bangkok, Thailand. Association for Computational Linguistics.

\bibitem[{D'Arcy et~al.(2024)D'Arcy, Hope, Birnbaum, and Downey}]{d2024marg}
Mike D'Arcy, Tom Hope, Larry Birnbaum, and Doug Downey. 2024.
\newblock \href {https://arxiv.org/abs/2401.04259v1} {{MARG}: Multi-agent review generation for scientific papers}.
\newblock \emph{arXiv preprint arXiv:2401.04259v1}.

\bibitem[{DeepSeek-AI et~al.(2024)DeepSeek-AI, Liu, Feng, Xue, Wang, Wu, Lu, Zhao, Deng, Zhang, Ruan, Dai, Guo, Yang, Chen, Ji, Li, Lin, Dai, Luo, Hao, Chen, Li, Zhang, Bao, Xu, Wang, Zhang, Ding, Xin, Gao, Li, Qu, Cai, Liang, Guo, Ni, Li, Wang, Chen, Chen, Yuan, Qiu, Li, Song, Dong, Hu, Gao, Guan, Huang, Yu, Wang, Zhang, Xu, Xia, Zhao, Wang, Zhang, Li, Wang, Zhang, Zhang, Tang, Li, Tian, Huang, Wang, Zhang, Wang, Zhu, Chen, Du, Chen, Jin, Ge, Zhang, Pan, Wang, Xu, Zhang, Chen, Li, Lu, Zhou, Chen, Wu, Ye, Ye, Ma, Wang, Zhou, Yu, Zhou, Pan, Wang, Yun, Pei, Sun, Xiao, Zeng, Zhao, An, Liu, Liang, Gao, Yu, Zhang, Li, Jin, Wang, Bi, Liu, Wang, Shen, Chen, Zhang, Chen, Nie, Sun, Wang, Cheng, Liu, Xie, Liu, Yu, Song, Shan, Zhou, Yang, Li, Su, Lin, Li, Wang, Wei, Zhu, Zhang, Xu, Xu, Huang, Li, Zhao, Sun, Li, Wang, Yu, Zheng, Zhang, Shi, Xiong, He, Tang, Piao, Wang, Tan, Ma, Liu, Guo, Wu, Ou, Zhu, Wang, Gong, Zou, He, Zha, Xiong, Ma, Yan, Luo, You, Liu, Zhou, Wu, Ren, Ren, Sha, Fu, Xu, Huang, Zhang, Xie, Zhang, Hao,
  Gou, Ma, Yan, Shao, Xu, Wu, Zhang, Li, Gu, Zhu, Liu, Li, Xie, Song, Gao, and Pan}]{liu2024deepseek}
DeepSeek-AI, Aixin Liu, Bei Feng, Bing Xue, Bingxuan Wang, Bochao Wu, Chengda Lu, Chenggang Zhao, Chengqi Deng, Chenyu Zhang, Chong Ruan, Damai Dai, Daya Guo, Dejian Yang, Deli Chen, Dongjie Ji, Erhang Li, Fangyun Lin, Fucong Dai, Fuli Luo, Guangbo Hao, Guanting Chen, Guowei Li, H.~Zhang, Han Bao, Hanwei Xu, Haocheng Wang, Haowei Zhang, Honghui Ding, Huajian Xin, Huazuo Gao, Hui Li, Hui Qu, J.~L. Cai, Jian Liang, Jianzhong Guo, Jiaqi Ni, Jiashi Li, Jiawei Wang, Jin Chen, Jingchang Chen, Jingyang Yuan, Junjie Qiu, Junlong Li, Junxiao Song, Kai Dong, Kai Hu, Kaige Gao, Kang Guan, Kexin Huang, Kuai Yu, Lean Wang, Lecong Zhang, Lei Xu, Leyi Xia, Liang Zhao, Litong Wang, Liyue Zhang, Meng Li, Miaojun Wang, Mingchuan Zhang, Minghua Zhang, Minghui Tang, Mingming Li, Ning Tian, Panpan Huang, Peiyi Wang, Peng Zhang, Qiancheng Wang, Qihao Zhu, Qinyu Chen, Qiushi Du, R.~J. Chen, R.~L. Jin, Ruiqi Ge, Ruisong Zhang, Ruizhe Pan, Runji Wang, Runxin Xu, Ruoyu Zhang, Ruyi Chen, S.~S. Li, Shanghao Lu, Shangyan Zhou, Shanhuang
  Chen, Shaoqing Wu, Shengfeng Ye, Shengfeng Ye, Shirong Ma, Shiyu Wang, Shuang Zhou, Shuiping Yu, Shunfeng Zhou, Shuting Pan, T.~Wang, Tao Yun, Tian Pei, Tianyu Sun, W.~L. Xiao, Wangding Zeng, Wanjia Zhao, Wei An, Wen Liu, Wenfeng Liang, Wenjun Gao, Wenqin Yu, Wentao Zhang, X.~Q. Li, Xiangyue Jin, Xianzu Wang, Xiao Bi, Xiaodong Liu, Xiaohan Wang, Xiaojin Shen, Xiaokang Chen, Xiaokang Zhang, Xiaosha Chen, Xiaotao Nie, Xiaowen Sun, Xiaoxiang Wang, Xin Cheng, Xin Liu, Xin Xie, Xingchao Liu, Xingkai Yu, Xinnan Song, Xinxia Shan, Xinyi Zhou, Xinyu Yang, Xinyuan Li, Xuecheng Su, Xuheng Lin, Y.~K. Li, Y.~Q. Wang, Y.~X. Wei, Y.~X. Zhu, Yang Zhang, Yanhong Xu, Yanhong Xu, Yanping Huang, Yao Li, Yao Zhao, Yaofeng Sun, Yaohui Li, Yaohui Wang, Yi~Yu, Yi~Zheng, Yichao Zhang, Yifan Shi, Yiliang Xiong, Ying He, Ying Tang, Yishi Piao, Yisong Wang, Yixuan Tan, Yiyang Ma, Yiyuan Liu, Yongqiang Guo, Yu~Wu, Yuan Ou, Yuchen Zhu, Yuduan Wang, Yue Gong, Yuheng Zou, Yujia He, Yukun Zha, Yunfan Xiong, Yunxian Ma, Yuting Yan, Yuxiang
  Luo, Yuxiang You, Yuxuan Liu, Yuyang Zhou, Z.~F. Wu, Z.~Z. Ren, Zehui Ren, Zhangli Sha, Zhe Fu, Zhean Xu, Zhen Huang, Zhen Zhang, Zhenda Xie, Zhengyan Zhang, Zhewen Hao, Zhibin Gou, Zhicheng Ma, Zhigang Yan, Zhihong Shao, Zhipeng Xu, Zhiyu Wu, Zhongyu Zhang, Zhuoshu Li, Zihui Gu, Zijia Zhu, Zijun Liu, Zilin Li, Ziwei Xie, Ziyang Song, Ziyi Gao, and Zizheng Pan. 2024.
\newblock \href {https://arxiv.org/abs/2412.19437v2} {Deepseek-v3 technical report}.
\newblock \emph{arXiv preprint arXiv:2412.19437v2}.

\bibitem[{Du et~al.(2024)Du, Wang, Zhao, Deng, Liu, Lou, Zou, Narayanan~Venkit, Zhang, Srinath, Zhang, Gupta, Li, Li, Wang, Liu, Liu, Gao, Xia, Xing, Jiayang, Wang, Su, Shah, Guo, Gu, Li, Wei, Wang, Cheng, Ranathunga, Fang, Fu, Liu, Huang, Blanco, Cao, Zhang, Yu, and Yin}]{du-etal-2024-llms}
Jiangshu Du, Yibo Wang, Wenting Zhao, Zhongfen Deng, Shuaiqi Liu, Renze Lou, Henry~Peng Zou, Pranav Narayanan~Venkit, Nan Zhang, Mukund Srinath, Haoran~Ranran Zhang, Vipul Gupta, Yinghui Li, Tao Li, Fei Wang, Qin Liu, Tianlin Liu, Pengzhi Gao, Congying Xia, Chen Xing, Cheng Jiayang, Zhaowei Wang, Ying Su, Raj~Sanjay Shah, Ruohao Guo, Jing Gu, Haoran Li, Kangda Wei, Zihao Wang, Lu~Cheng, Surangika Ranathunga, Meng Fang, Jie Fu, Fei Liu, Ruihong Huang, Eduardo Blanco, Yixin Cao, Rui Zhang, Philip~S. Yu, and Wenpeng Yin. 2024.
\newblock \href {https://doi.org/10.18653/v1/2024.emnlp-main.292} {{LLM}s assist {NLP} researchers: Critique paper (meta-)reviewing}.
\newblock In \emph{Proceedings of the 2024 Conference on Empirical Methods in Natural Language Processing}, pages 5081--5099, Miami, Florida, USA. Association for Computational Linguistics.

\bibitem[{Dutta et~al.(2025)Dutta, Kaufmann, Glava{\v{s}}, Habernal, Kersting, Kreuter, Mezini, Gurevych, H{\"u}llermeier, and Schuetze}]{dutta2025problemsolvinghumanaipreferencebased}
Subhabrata Dutta, Timo Kaufmann, Goran Glava{\v{s}}, Ivan Habernal, Kristian Kersting, Frauke Kreuter, Mira Mezini, Iryna Gurevych, Eyke H{\"u}llermeier, and Hinrich Schuetze. 2025.
\newblock Problem solving through human--{AI} preference-based cooperation.
\newblock \emph{Computational Linguistics}, pages 1--36.

\bibitem[{Dycke et~al.(2023{\natexlab{a}})Dycke, Kuznetsov, and Gurevych}]{dycke-etal-2023-nlpeer}
Nils Dycke, Ilia Kuznetsov, and Iryna Gurevych. 2023{\natexlab{a}}.
\newblock \href {https://doi.org/10.18653/v1/2023.acl-long.277} {{NLP}eer: A unified resource for the computational study of peer review}.
\newblock In \emph{Proceedings of the 61st Annual Meeting of the Association for Computational Linguistics (Volume 1: Long Papers)}, pages 5049--5073, Toronto, Canada. Association for Computational Linguistics.

\bibitem[{Dycke et~al.(2023{\natexlab{b}})Dycke, Kuznetsov, and Gurevych}]{dycke2023overview}
Nils Dycke, Ilia Kuznetsov, and Iryna Gurevych. 2023{\natexlab{b}}.
\newblock \href {https://doi.org/10.18653/v1/2023.argmining-1.21} {Overview of {P}rag{T}ag-2023: Low-resource multi-domain pragmatic tagging of peer reviews}.
\newblock In \emph{Proceedings of the 10th Workshop on Argument Mining}, pages 187--196, Singapore. Association for Computational Linguistics.

\bibitem[{Dycke et~al.(2025)Dycke, Ze{\v{c}}evi{\'c}, Kuznetsov, Suess, Kersting, and Gurevych}]{dycke2025strictastructuredreasoningcritical}
Nils Dycke, Matej Ze{\v{c}}evi{\'c}, Ilia Kuznetsov, Beatrix Suess, Kristian Kersting, and Iryna Gurevych. 2025.
\newblock \href {https://aclanthology.org/2025.acl-long.1107/} {{STRICTA}: Structured reasoning in critical text assessment for peer review and beyond}.
\newblock In \emph{Proceedings of the 63rd Annual Meeting of the Association for Computational Linguistics (Volume 1: Long Papers)}, pages 22687--22727, Vienna, Austria. Association for Computational Linguistics.

\bibitem[{Fang et~al.(2024)Fang, Xu, Tan, Hu, Zhang, Qi, Sengamedu, and Faloutsos}]{fang2024large}
Xi~Fang, Weijie Xu, Fiona~Anting Tan, Ziqing Hu, Jiani Zhang, Yanjun Qi, Srinivasan~H. Sengamedu, and Christos Faloutsos. 2024.
\newblock \href {https://openreview.net/forum?id=IZnrCGF9WI} {Large language models ({LLM}s) on tabular data: Prediction, generation, and understanding - a survey}.
\newblock \emph{Transactions on Machine Learning Research}.

\bibitem[{Feinstein and Cicchetti(1990)}]{feinstein1990high}
Alvan~R. Feinstein and Domenic~V. Cicchetti. 1990.
\newblock High agreement but low kappa: I. the problems of two paradoxes.
\newblock \emph{Journal of clinical epidemiology}, 43(6):543--549.

\bibitem[{Gallegos et~al.(2024)Gallegos, Rossi, Barrow, Tanjim, Kim, Dernoncourt, Yu, Zhang, and Ahmed}]{gallegos-etal-2024-bias}
Isabel~O. Gallegos, Ryan~A. Rossi, Joe Barrow, Md~Mehrab Tanjim, Sungchul Kim, Franck Dernoncourt, Tong Yu, Ruiyi Zhang, and Nesreen~K. Ahmed. 2024.
\newblock \href {https://doi.org/10.1162/coli_a_00524} {Bias and fairness in large language models: A survey}.
\newblock \emph{Computational Linguistics}, 50(3):1097--1179.

\bibitem[{Gao et~al.(2024)Gao, Brantley, and Joachims}]{gao2024reviewer2}
Zhaolin Gao, Kiant{\'e} Brantley, and Thorsten Joachims. 2024.
\newblock \href {https://arxiv.org/abs/2402.10886v2} {Reviewer2: Optimizing review generation through prompt generation}.
\newblock \emph{arXiv preprint arXiv:2402.10886v2}.

\bibitem[{Gat et~al.(2024)Gat, Calderon, Feder, Chapanin, Sharma, and Reichart}]{gat2024counterfactuals}
Yair~Ori Gat, Nitay Calderon, Amir Feder, Alexander Chapanin, Amit Sharma, and Roi Reichart. 2024.
\newblock \href {https://openreview.net/forum?id=UMfcdRIotC} {Faithful explanations of black-box {NLP} models using {LLM}-generated counterfactuals}.
\newblock In \emph{The Twelfth International Conference on Learning Representations, {ICLR} 2024, Vienna, Austria, May 7-11, 2024}. OpenReview.net.

\bibitem[{Goyal et~al.(2019)Goyal, Feder, Shalit, and Kim}]{goyal2019explaining}
Yash Goyal, Amir Feder, Uri Shalit, and Been Kim. 2019.
\newblock \href {https://arxiv.org/abs/1907.07165v2} {Explaining classifiers with causal concept effect ({CaCE})}.
\newblock \emph{arXiv preprint arXiv:1907.07165v2}.

\bibitem[{Hsieh et~al.(2024)Hsieh, Sun, Kriman, Acharya, Rekesh, Jia, and Ginsburg}]{hsieh2024ruler}
Cheng-Ping Hsieh, Simeng Sun, Samuel Kriman, Shantanu Acharya, Dima Rekesh, Fei Jia, and Boris Ginsburg. 2024.
\newblock \href {https://openreview.net/forum?id=kIoBbc76Sy} {{RULER}: What{\textquoteright}s the real context size of your long-context language models?}
\newblock In \emph{First Conference on Language Modeling}.

\bibitem[{Huang et~al.(2025)Huang, Chan, Fung, Qiu, Zhou, Joty, Chang, and Ji}]{llmchartunderstanding2025}
Kung-Hsiang Huang, Hou~Pong Chan, May Fung, Haoyi Qiu, Mingyang Zhou, Shafiq Joty, Shih-Fu Chang, and Heng Ji. 2025.
\newblock \href {https://doi.org/10.1109/TKDE.2024.3513320} {From pixels to insights: A survey on automatic chart understanding in the era of large foundation models}.
\newblock \emph{IEEE Transactions on Knowledge and Data Engineering}, 37(5):2550--2568.

\bibitem[{Idahl and Ahmadi(2025)}]{idahl2024openreviewer}
Maximilian Idahl and Zahra Ahmadi. 2025.
\newblock \href {https://doi.org/10.18653/v1/2025.naacl-demo.44} {{O}pen{R}eviewer: A specialized large language model for generating critical scientific paper reviews}.
\newblock In \emph{Proceedings of the 2025 Conference of the Nations of the Americas Chapter of the Association for Computational Linguistics: Human Language Technologies (System Demonstrations)}, pages 550--562, Albuquerque, New Mexico. Association for Computational Linguistics.

\bibitem[{Jefferson et~al.(2002)Jefferson, Wager, and Davidoff}]{jefferson2002prquality}
Tom Jefferson, Elizabeth Wager, and Frank Davidoff. 2002.
\newblock \href {https://doi.org/10.1001/jama.287.21.2786} {Measuring the quality of editorial peer review}.
\newblock \emph{JAMA}, 287(21):2786--2790.

\bibitem[{Ji et~al.(2023)Ji, Lee, Frieske, Yu, Su, Xu, Ishii, Bang, Madotto, and Fung}]{ji2023hallucination}
Ziwei Ji, Nayeon Lee, Rita Frieske, Tiezheng Yu, Dan Su, Yan Xu, Etsuko Ishii, Ye~Jin Bang, Andrea Madotto, and Pascale Fung. 2023.
\newblock \href {https://doi.org/10.1145/3571730} {Survey of hallucination in natural language generation}.
\newblock \emph{ACM Computing Surveys}, 55(12).

\bibitem[{Kirtani et~al.(2025)Kirtani, Garg, Prasad, Singhal, Mandal, and Kumar}]{kirtani2025revieweval}
Chhavi Kirtani, Madhav~Krishan Garg, Tejash Prasad, Tanmay Singhal, Murari Mandal, and Dhruv Kumar. 2025.
\newblock \href {https://arxiv.org/abs/2502.11736v3} {{R}eview{E}val: An evaluation framework for {AI}-generated reviews}.
\newblock \emph{arXiv preprint arXiv:2502.11736v3}.

\bibitem[{Kuznetsov et~al.(2024)Kuznetsov, Afzal, Dercksen, Dycke, Goldberg, Hope, Hovy, Kummerfeld, Lauscher, Leyton-Brown, Lu, Mausam, Mieskes, Névéol, Pruthi, Qu, Schwartz, Smith, Solorio, Wang, Zhu, Rogers, Shah, and Gurevych}]{kuznetsov2024can}
Ilia Kuznetsov, Osama~Mohammed Afzal, Koen Dercksen, Nils Dycke, Alexander Goldberg, Tom Hope, Dirk Hovy, Jonathan~K. Kummerfeld, Anne Lauscher, Kevin Leyton-Brown, Sheng Lu, Mausam, Margot Mieskes, Aurélie Névéol, Danish Pruthi, Lizhen Qu, Roy Schwartz, Noah~A. Smith, Thamar Solorio, Jingyan Wang, Xiaodan Zhu, Anna Rogers, Nihar~B. Shah, and Iryna Gurevych. 2024.
\newblock \href {https://arxiv.org/abs/2405.06563v1} {What can natural language processing do for peer review?}
\newblock \emph{arXiv preprint arXiv:2405.06563v1}.

\bibitem[{Li et~al.(2025)Li, Li, Hu, Gao, and Wan}]{li2025aspect}
Jiatao Li, Yanheng Li, Xinyu Hu, Mingqi Gao, and Xiaojun Wan. 2025.
\newblock \href {https://arxiv.org/abs/2502.12510v1} {Aspect-guided multi-level perturbation analysis of large language models in automated peer review}.
\newblock \emph{arXiv preprint arXiv:2502.12510v1}.

\bibitem[{Li et~al.(2024)Li, Xu, Miao, Zhou, and Qian}]{li2025llmcounter}
Yongqi Li, Mayi Xu, Xin Miao, Shen Zhou, and Tieyun Qian. 2024.
\newblock \href {https://aclanthology.org/2024.lrec-main.1156} {Prompting large language models for counterfactual generation: An empirical study}.
\newblock In \emph{Proceedings of the 2024 Joint International Conference on Computational Linguistics, Language Resources and Evaluation, {LREC/COLING} 2024, 20-25 May, 2024, Torino, Italy}, pages 13201--13221. {ELRA} and {ICCL}.

\bibitem[{Liang et~al.(2024{\natexlab{a}})Liang, Izzo, Zhang, Lepp, Cao, Zhao, Chen, Ye, Liu, Huang, Mcfarland, and Zou}]{pmlr-v235-liang24b}
Weixin Liang, Zachary Izzo, Yaohui Zhang, Haley Lepp, Hancheng Cao, Xuandong Zhao, Lingjiao Chen, Haotian Ye, Sheng Liu, Zhi Huang, Daniel Mcfarland, and James~Y. Zou. 2024{\natexlab{a}}.
\newblock \href {https://proceedings.mlr.press/v235/liang24b.html} {Monitoring {AI}-modified content at scale: A case study on the impact of {C}hat{GPT} on {AI} conference peer reviews}.
\newblock In \emph{Proceedings of the 41st International Conference on Machine Learning}, volume 235 of \emph{Proceedings of Machine Learning Research}, pages 29575--29620. PMLR.

\bibitem[{Liang et~al.(2024{\natexlab{b}})Liang, Zhang, Cao, Wang, Ding, Yang, Vodrahalli, He, Smith, Yin, McFarland, and Zou}]{liang2024can}
Weixin Liang, Yuhui Zhang, Hancheng Cao, Binglu Wang, Daisy~Yi Ding, Xinyu Yang, Kailas Vodrahalli, Siyu He, Daniel~Scott Smith, Yian Yin, Daniel~A. McFarland, and James Zou. 2024{\natexlab{b}}.
\newblock \href {https://doi.org/10.1056/AIoa2400196} {Can large language models provide useful feedback on research papers? {A} large-scale empirical analysis}.
\newblock \emph{NEJM AI}, 1(8):AIoa2400196.

\bibitem[{Lin et~al.(2024)Lin, Chiang, and Lee}]{lin-etal-2024-advancing}
Guan-Ting Lin, Cheng-Han Chiang, and Hung-yi Lee. 2024.
\newblock \href {https://doi.org/10.18653/v1/2024.acl-long.358} {Advancing large language models to capture varied speaking styles and respond properly in spoken conversations}.
\newblock In \emph{Proceedings of the 62nd Annual Meeting of the Association for Computational Linguistics (Volume 1: Long Papers)}, pages 6626--6642, Bangkok, Thailand. Association for Computational Linguistics.

\bibitem[{Lindstrom and Bates(1988)}]{lindstrom1988newton}
Mary~J. Lindstrom and Douglas~M. Bates. 1988.
\newblock {N}ewton—{R}aphson and {EM} algorithms for linear mixed-effects models for repeated-measures data.
\newblock \emph{Journal of the American Statistical Association}, 83(404):1014--1022.

\bibitem[{Liu and Shah(2023)}]{liu2023reviewergpt}
Ryan Liu and Nihar~B Shah. 2023.
\newblock \href {https://arxiv.org/abs/2306.00622v1} {Reviewer{GPT}? {A}n exploratory study on using large language models for paper reviewing}.
\newblock \emph{arXiv preprint arXiv:2306.00622v1}.

\bibitem[{Lou et~al.(2025)Lou, Xu, Wang, Du, Kamoi, Lu, Xie, Sun, Zhang, Ahn, Fang, Zou, Ma, Li, Zhang, Xia, Huang, and Yin}]{lou2024aaar}
Renze Lou, Hanzi Xu, Sijia Wang, Jiangshu Du, Ryo Kamoi, Xiaoxin Lu, Jian Xie, Yuxuan Sun, Yusen Zhang, Jihyun~Janice Ahn, Hongchao Fang, Zhuoyang Zou, Wenchao Ma, Xi~Li, Kai Zhang, Congying Xia, Lifu Huang, and Wenpeng Yin. 2025.
\newblock \href {https://openreview.net/forum?id=RHAWcjIyl2} {{AAAR}-1.0: Assessing {AI}{\textquoteright}s potential to assist research}.
\newblock In \emph{Forty-second International Conference on Machine Learning}.

\bibitem[{Lu et~al.(2025)Lu, Kuznetsov, and Gurevych}]{lu2025identifying}
Sheng Lu, Ilia Kuznetsov, and Iryna Gurevych. 2025.
\newblock \href {https://doi.org/10.18653/v1/2025.findings-emnlp.326} {Identifying aspects in peer reviews}.
\newblock In \emph{Findings of the Association for Computational Linguistics: EMNLP 2025}, pages 6145--6167, Suzhou, China. Association for Computational Linguistics.

\bibitem[{Madaan et~al.(2023)Madaan, Tandon, Gupta, Hallinan, Gao, Wiegreffe, Alon, Dziri, Prabhumoye, Yang, Gupta, Majumder, Hermann, Welleck, Yazdanbakhsh, and Clark}]{madaan2023self}
Aman Madaan, Niket Tandon, Prakhar Gupta, Skyler Hallinan, Luyu Gao, Sarah Wiegreffe, Uri Alon, Nouha Dziri, Shrimai Prabhumoye, Yiming Yang, Shashank Gupta, Bodhisattwa~Prasad Majumder, Katherine Hermann, Sean Welleck, Amir Yazdanbakhsh, and Peter Clark. 2023.
\newblock \href {https://proceedings.neurips.cc/paper_files/paper/2023/file/91edff07232fb1b55a505a9e9f6c0ff3-Paper-Conference.pdf} {Self-refine: Iterative refinement with self-feedback}.
\newblock In \emph{Advances in Neural Information Processing Systems}, volume~36, pages 46534--46594. Curran Associates, Inc.

\bibitem[{McCook(2006)}]{mccook2006peer}
Alison McCook. 2006.
\newblock Is peer review broken? {S}ubmissions are up, reviewers are overtaxed, and authors are lodging complaint after complaint about the process at top-tier journals. {W}hat's wrong with peer review?
\newblock \emph{The scientist}, 20(2):26--35.

\bibitem[{Molnar(2025)}]{molnar2025}
Christoph Molnar. 2025.
\newblock \href {https://christophm.github.io/interpretable-ml-book} {\emph{Interpretable Machine Learning}}, 3rd edition.

\bibitem[{Mu and Li(2024)}]{mu-li-2024-causal}
Feiteng Mu and Wenjie Li. 2024.
\newblock \href {https://doi.org/10.18653/v1/2024.acl-long.354} {A causal approach for counterfactual reasoning in narratives}.
\newblock In \emph{Proceedings of the 62nd Annual Meeting of the Association for Computational Linguistics (Volume 1: Long Papers)}, pages 6556--6569, Bangkok, Thailand. Association for Computational Linguistics.

\bibitem[{OpenAI et~al.(2024)OpenAI, Achiam, Adler, Agarwal, Ahmad, Akkaya, Aleman, Almeida, Altenschmidt, Altman, Anadkat, Avila, Babuschkin, Balaji, Balcom, Baltescu, Bao, Bavarian, Belgum, Bello, Berdine, Bernadett-Shapiro, Berner, Bogdonoff, Boiko, Boyd, Brakman, Brockman, Brooks, Brundage, Button, Cai, Campbell, Cann, Carey, Carlson, Carmichael, Chan, Chang, Chantzis, Chen, Chen, Chen, Chen, Chen, Chess, Cho, Chu, Chung, Cummings, Currier, Dai, Decareaux, Degry, Deutsch, Deville, Dhar, Dohan, Dowling, Dunning, Ecoffet, Eleti, Eloundou, Farhi, Fedus, Felix, Fishman, Forte, Fulford, Gao, Georges, Gibson, Goel, Gogineni, Goh, Gontijo-Lopes, Gordon, Grafstein, Gray, Greene, Gross, Gu, Guo, Hallacy, Han, Harris, He, Heaton, Heidecke, Hesse, Hickey, Hickey, Hoeschele, Houghton, Hsu, Hu, Hu, Huizinga, Jain, Jain, Jang, Jiang, Jiang, Jin, Jin, Jomoto, Jonn, Jun, Kaftan, Łukasz Kaiser, Kamali, Kanitscheider, Keskar, Khan, Kilpatrick, Kim, Kim, Kim, Kirchner, Kiros, Knight, Kokotajlo, Łukasz Kondraciuk,
  Kondrich, Konstantinidis, Kosic, Krueger, Kuo, Lampe, Lan, Lee, Leike, Leung, Levy, Li, Lim, Lin, Lin, Litwin, Lopez, Lowe, Lue, Makanju, Malfacini, Manning, Markov, Markovski, Martin, Mayer, Mayne, McGrew, McKinney, McLeavey, McMillan, McNeil, Medina, Mehta, Menick, Metz, Mishchenko, Mishkin, Monaco, Morikawa, Mossing, Mu, Murati, Murk, Mély, Nair, Nakano, Nayak, Neelakantan, Ngo, Noh, Ouyang, O'Keefe, Pachocki, Paino, Palermo, Pantuliano, Parascandolo, Parish, Parparita, Passos, Pavlov, Peng, Perelman, de~Avila Belbute~Peres, Petrov, de~Oliveira~Pinto, Michael, Pokorny, Pokrass, Pong, Powell, Power, Power, Proehl, Puri, Radford, Rae, Ramesh, Raymond, Real, Rimbach, Ross, Rotsted, Roussez, Ryder, Saltarelli, Sanders, Santurkar, Sastry, Schmidt, Schnurr, Schulman, Selsam, Sheppard, Sherbakov, Shieh, Shoker, Shyam, Sidor, Sigler, Simens, Sitkin, Slama, Sohl, Sokolowsky, Song, Staudacher, Such, Summers, Sutskever, Tang, Tezak, Thompson, Tillet, Tootoonchian, Tseng, Tuggle, Turley, Tworek, Uribe, Vallone,
  Vijayvergiya, Voss, Wainwright, Wang, Wang, Wang, Ward, Wei, Weinmann, Welihinda, Welinder, Weng, Weng, Wiethoff, Willner, Winter, Wolrich, Wong, Workman, Wu, Wu, Wu, Xiao, Xu, Yoo, Yu, Yuan, Zaremba, Zellers, Zhang, Zhang, Zhao, Zheng, Zhuang, Zhuk, and Zoph}]{openai2024gpt4technicalreport}
OpenAI, Josh Achiam, Steven Adler, Sandhini Agarwal, Lama Ahmad, Ilge Akkaya, Florencia~Leoni Aleman, Diogo Almeida, Janko Altenschmidt, Sam Altman, Shyamal Anadkat, Red Avila, Igor Babuschkin, Suchir Balaji, Valerie Balcom, Paul Baltescu, Haiming Bao, Mohammad Bavarian, Jeff Belgum, Irwan Bello, Jake Berdine, Gabriel Bernadett-Shapiro, Christopher Berner, Lenny Bogdonoff, Oleg Boiko, Madelaine Boyd, Anna-Luisa Brakman, Greg Brockman, Tim Brooks, Miles Brundage, Kevin Button, Trevor Cai, Rosie Campbell, Andrew Cann, Brittany Carey, Chelsea Carlson, Rory Carmichael, Brooke Chan, Che Chang, Fotis Chantzis, Derek Chen, Sully Chen, Ruby Chen, Jason Chen, Mark Chen, Ben Chess, Chester Cho, Casey Chu, Hyung~Won Chung, Dave Cummings, Jeremiah Currier, Yunxing Dai, Cory Decareaux, Thomas Degry, Noah Deutsch, Damien Deville, Arka Dhar, David Dohan, Steve Dowling, Sheila Dunning, Adrien Ecoffet, Atty Eleti, Tyna Eloundou, David Farhi, Liam Fedus, Niko Felix, Simón~Posada Fishman, Juston Forte, Isabella Fulford, Leo
  Gao, Elie Georges, Christian Gibson, Vik Goel, Tarun Gogineni, Gabriel Goh, Rapha Gontijo-Lopes, Jonathan Gordon, Morgan Grafstein, Scott Gray, Ryan Greene, Joshua Gross, Shixiang~Shane Gu, Yufei Guo, Chris Hallacy, Jesse Han, Jeff Harris, Yuchen He, Mike Heaton, Johannes Heidecke, Chris Hesse, Alan Hickey, Wade Hickey, Peter Hoeschele, Brandon Houghton, Kenny Hsu, Shengli Hu, Xin Hu, Joost Huizinga, Shantanu Jain, Shawn Jain, Joanne Jang, Angela Jiang, Roger Jiang, Haozhun Jin, Denny Jin, Shino Jomoto, Billie Jonn, Heewoo Jun, Tomer Kaftan, Łukasz Kaiser, Ali Kamali, Ingmar Kanitscheider, Nitish~Shirish Keskar, Tabarak Khan, Logan Kilpatrick, Jong~Wook Kim, Christina Kim, Yongjik Kim, Jan~Hendrik Kirchner, Jamie Kiros, Matt Knight, Daniel Kokotajlo, Łukasz Kondraciuk, Andrew Kondrich, Aris Konstantinidis, Kyle Kosic, Gretchen Krueger, Vishal Kuo, Michael Lampe, Ikai Lan, Teddy Lee, Jan Leike, Jade Leung, Daniel Levy, Chak~Ming Li, Rachel Lim, Molly Lin, Stephanie Lin, Mateusz Litwin, Theresa Lopez, Ryan
  Lowe, Patricia Lue, Anna Makanju, Kim Malfacini, Sam Manning, Todor Markov, Yaniv Markovski, Bianca Martin, Katie Mayer, Andrew Mayne, Bob McGrew, Scott~Mayer McKinney, Christine McLeavey, Paul McMillan, Jake McNeil, David Medina, Aalok Mehta, Jacob Menick, Luke Metz, Andrey Mishchenko, Pamela Mishkin, Vinnie Monaco, Evan Morikawa, Daniel Mossing, Tong Mu, Mira Murati, Oleg Murk, David Mély, Ashvin Nair, Reiichiro Nakano, Rajeev Nayak, Arvind Neelakantan, Richard Ngo, Hyeonwoo Noh, Long Ouyang, Cullen O'Keefe, Jakub Pachocki, Alex Paino, Joe Palermo, Ashley Pantuliano, Giambattista Parascandolo, Joel Parish, Emy Parparita, Alex Passos, Mikhail Pavlov, Andrew Peng, Adam Perelman, Filipe de~Avila Belbute~Peres, Michael Petrov, Henrique~Ponde de~Oliveira~Pinto, Michael, Pokorny, Michelle Pokrass, Vitchyr~H. Pong, Tolly Powell, Alethea Power, Boris Power, Elizabeth Proehl, Raul Puri, Alec Radford, Jack Rae, Aditya Ramesh, Cameron Raymond, Francis Real, Kendra Rimbach, Carl Ross, Bob Rotsted, Henri Roussez,
  Nick Ryder, Mario Saltarelli, Ted Sanders, Shibani Santurkar, Girish Sastry, Heather Schmidt, David Schnurr, John Schulman, Daniel Selsam, Kyla Sheppard, Toki Sherbakov, Jessica Shieh, Sarah Shoker, Pranav Shyam, Szymon Sidor, Eric Sigler, Maddie Simens, Jordan Sitkin, Katarina Slama, Ian Sohl, Benjamin Sokolowsky, Yang Song, Natalie Staudacher, Felipe~Petroski Such, Natalie Summers, Ilya Sutskever, Jie Tang, Nikolas Tezak, Madeleine~B. Thompson, Phil Tillet, Amin Tootoonchian, Elizabeth Tseng, Preston Tuggle, Nick Turley, Jerry Tworek, Juan Felipe~Cerón Uribe, Andrea Vallone, Arun Vijayvergiya, Chelsea Voss, Carroll Wainwright, Justin~Jay Wang, Alvin Wang, Ben Wang, Jonathan Ward, Jason Wei, CJ~Weinmann, Akila Welihinda, Peter Welinder, Jiayi Weng, Lilian Weng, Matt Wiethoff, Dave Willner, Clemens Winter, Samuel Wolrich, Hannah Wong, Lauren Workman, Sherwin Wu, Jeff Wu, Michael Wu, Kai Xiao, Tao Xu, Sarah Yoo, Kevin Yu, Qiming Yuan, Wojciech Zaremba, Rowan Zellers, Chong Zhang, Marvin Zhang, Shengjia
  Zhao, Tianhao Zheng, Juntang Zhuang, William Zhuk, and Barret Zoph. 2024.
\newblock \href {http://arxiv.org/abs/2303.08774} {{GPT}-4 technical report}.
\newblock \emph{arXiv preprint arXiv:2303.08774v6}.

\bibitem[{Pearl(2000)}]{pearl2000models}
Judea Pearl. 2000.
\newblock \emph{Causality: {M}odels, {R}easoning and {I}nference}, 2 edition, volume~7.
\newblock Cambridge University Press.

\bibitem[{Phoenix and Taylor(2024)}]{phoenix2024prompt}
James Phoenix and Mike Taylor. 2024.
\newblock \emph{Prompt {E}ngineering for {G}enerative {AI}}.
\newblock O'Reilly Media, Inc.

\bibitem[{Rao et~al.(2023)Rao, Leung, and Miao}]{rao-etal-2023-chatgpt}
Haocong Rao, Cyril Leung, and Chunyan Miao. 2023.
\newblock \href {https://doi.org/10.18653/v1/2023.findings-emnlp.84} {Can {C}hat{GPT} assess human personalities? {A} general evaluation framework}.
\newblock In \emph{Findings of the Association for Computational Linguistics: EMNLP 2023}, pages 1184--1194, Singapore. Association for Computational Linguistics.

\bibitem[{Sahoo et~al.(2024)Sahoo, Singh, Saha, Jain, Mondal, and Chadha}]{sahoo2024systematic}
Pranab Sahoo, Ayush~Kumar Singh, Sriparna Saha, Vinija Jain, Samrat Mondal, and Aman Chadha. 2024.
\newblock \href {https://arxiv.org/abs/2402.07927v2} {A systematic survey of prompt engineering in large language models: {T}echniques and applications}.
\newblock \emph{arXiv preprint arXiv:2402.07927v2}.

\bibitem[{Sainz et~al.(2023)Sainz, Campos, Garc{\'i}a-Ferrero, Etxaniz, de~Lacalle, and Agirre}]{sainz-etal-2023-nlp}
Oscar Sainz, Jon Campos, Iker Garc{\'i}a-Ferrero, Julen Etxaniz, Oier~Lopez de~Lacalle, and Eneko Agirre. 2023.
\newblock \href {https://doi.org/10.18653/v1/2023.findings-emnlp.722} {{NLP} evaluation in trouble: On the need to measure {LLM} data contamination for each benchmark}.
\newblock In \emph{Findings of the Association for Computational Linguistics: EMNLP 2023}, pages 10776--10787, Singapore. Association for Computational Linguistics.

\bibitem[{Sclar et~al.(2024)Sclar, Choi, Tsvetkov, and Suhr}]{sclar2024quantifying}
Melanie Sclar, Yejin Choi, Yulia Tsvetkov, and Alane Suhr. 2024.
\newblock \href {https://openreview.net/forum?id=RIu5lyNXjT} {Quantifying language models' sensitivity to spurious features in prompt design or: How i learned to start worrying about prompt formatting}.
\newblock In \emph{The Twelfth International Conference on Learning Representations}.

\bibitem[{Sculley et~al.(2018)Sculley, Snoek, and Wiltschko}]{sculley2018avoiding}
D~Sculley, Jasper Snoek, and Alex Wiltschko. 2018.
\newblock \href {https://arxiv.org/abs/1901.06246v1} {Avoiding a tragedy of the commons in the peer review process}.
\newblock \emph{arXiv preprint arXiv:1901.06246v1}.

\bibitem[{Shin et~al.(2025)Shin, Tang, Lee, Kim, Lim, Cho, Hong, Lee, and Kim}]{shin2025mind}
Hyungyu Shin, Jingyu Tang, Yoonjoo Lee, Nayoung Kim, Hyunseung Lim, Ji~Yong Cho, Hwajung Hong, Moontae Lee, and Juho Kim. 2025.
\newblock \href {https://arxiv.org/abs/2502.17086v4} {Mind the blind spots: {A} focus-level evaluation framework for {LLM} reviews}.
\newblock \emph{arXiv preprint arXiv:2502.17086v4}.

\bibitem[{Son et~al.(2025)Son, Hong, Fan, Nam, Ko, Lim, Song, Choi, Paulo, Yu, and Biderman}]{son2025ai}
Guijin Son, Jiwoo Hong, Honglu Fan, Heejeong Nam, Hyunwoo Ko, Seungwon Lim, Jinyeop Song, Jinha Choi, Gonçalo Paulo, Youngjae Yu, and Stella Biderman. 2025.
\newblock \href {https://arxiv.org/abs/2505.11855v1} {When {AI} co-scientists fail: {SPOT}-a benchmark for automated verification of scientific research}.
\newblock \emph{arXiv preprint arXiv:2505.11855v1}.

\bibitem[{Teufel et~al.(2009)Teufel, Siddharthan, and Batchelor}]{teufel-etal-2009-towards}
Simone Teufel, Advaith Siddharthan, and Colin Batchelor. 2009.
\newblock \href {https://aclanthology.org/D09-1155/} {Towards domain-independent argumentative zoning: Evidence from chemistry and computational linguistics}.
\newblock In \emph{Proceedings of the 2009 Conference on Empirical Methods in Natural Language Processing}, pages 1493--1502, Singapore. Association for Computational Linguistics.

\bibitem[{Touvron et~al.(2023)Touvron, Martin, Stone, Albert, Almahairi, Babaei, Bashlykov, Batra, Bhargava, Bhosale, Bikel, Blecher, Canton{-}Ferrer, Chen, Cucurull, Esiobu, Fernandes, Fu, Fu, Fuller, Gao, Goswami, Goyal, Hartshorn, Hosseini, Hou, Inan, Kardas, Kerkez, Khabsa, Kloumann, Korenev, Koura, Lachaux, Lavril, Lee, Liskovich, Lu, Mao, Martinet, Mihaylov, Mishra, Molybog, Nie, Poulton, Reizenstein, Rungta, Saladi, Schelten, Silva, Smith, Subramanian, Tan, Tang, Taylor, Williams, Kuan, Xu, Yan, Zarov, Zhang, Fan, Kambadur, Narang, Rodriguez, Stojnic, Edunov, and Scialom}]{llama2}
Hugo Touvron, Louis Martin, Kevin Stone, Peter Albert, Amjad Almahairi, Yasmine Babaei, Nikolay Bashlykov, Soumya Batra, Prajjwal Bhargava, Shruti Bhosale, Dan Bikel, Lukas Blecher, Cristian Canton{-}Ferrer, Moya Chen, Guillem Cucurull, David Esiobu, Jude Fernandes, Jeremy Fu, Wenyin Fu, Brian Fuller, Cynthia Gao, Vedanuj Goswami, Naman Goyal, Anthony Hartshorn, Saghar Hosseini, Rui Hou, Hakan Inan, Marcin Kardas, Viktor Kerkez, Madian Khabsa, Isabel Kloumann, Artem Korenev, Punit~Singh Koura, Marie{-}Anne Lachaux, Thibaut Lavril, Jenya Lee, Diana Liskovich, Yinghai Lu, Yuning Mao, Xavier Martinet, Todor Mihaylov, Pushkar Mishra, Igor Molybog, Yixin Nie, Andrew Poulton, Jeremy Reizenstein, Rashi Rungta, Kalyan Saladi, Alan Schelten, Ruan Silva, Eric~Michael Smith, Ranjan Subramanian, Xiaoqing~Ellen Tan, Binh Tang, Ross Taylor, Adina Williams, Jian~Xiang Kuan, Puxin Xu, Zheng Yan, Iliyan Zarov, Yuchen Zhang, Angela Fan, Melanie Kambadur, Sharan Narang, Aur{\'{e}}lien Rodriguez, Robert Stojnic, Sergey Edunov,
  and Thomas Scialom. 2023.
\newblock \href {https://doi.org/10.48550/ARXIV.2307.09288} {Llama 2: Open foundation and fine-tuned chat models}.
\newblock \emph{CoRR}, abs/2307.09288.

\bibitem[{Tyser et~al.(2024)Tyser, Segev, Longhitano, Zhang, Meeks, Lee, Garg, Belsten, Shporer, Udell, Te'eni, and Drori}]{tyser2024ai}
Keith Tyser, Ben Segev, Gaston Longhitano, Xin-Yu Zhang, Zachary Meeks, Jason Lee, Uday Garg, Nicholas Belsten, Avi Shporer, Madeleine Udell, Dov Te'eni, and Iddo Drori. 2024.
\newblock \href {https://arxiv.org/abs/2408.10365v1} {{AI}-driven review systems: {E}valuating {LLM}s in scalable and bias-aware academic reviews}.
\newblock \emph{arXiv preprint arXiv:2408.10365v1}.

\bibitem[{Waltman et~al.(2023)Waltman, Kaltenbrunner, Pinfield, and Woods}]{waltman2023improve}
Ludo Waltman, Wolfgang Kaltenbrunner, Stephen Pinfield, and Helen~Buckley Woods. 2023.
\newblock How to improve scientific peer review: Four schools of thought.
\newblock \emph{Learned Publishing}, 36(3):334--347.

\bibitem[{Wang et~al.(2024{\natexlab{a}})Wang, Qiu, Yue, Guo, Zeng, Feng, and Shen}]{wang-etal-2024-survey}
Yongjie Wang, Xiaoqi Qiu, Yu~Yue, Xu~Guo, Zhiwei Zeng, Yuhong Feng, and Zhiqi Shen. 2024{\natexlab{a}}.
\newblock \href {https://doi.org/10.18653/v1/2024.findings-emnlp.276} {A survey on natural language counterfactual generation}.
\newblock In \emph{Findings of the Association for Computational Linguistics: EMNLP 2024}, pages 4798--4818, Miami, Florida, USA. Association for Computational Linguistics.

\bibitem[{Wang et~al.(2024{\natexlab{b}})Wang, Zhang, and Du}]{wang2024beyond}
Ziao Wang, Xiaofeng Zhang, and Hongwei Du. 2024{\natexlab{b}}.
\newblock Beyond what if: Advancing counterfactual text generation with structural causal modeling.
\newblock In \emph{Proceedings of the Thirty-Third International Joint Conference on Artificial Intelligence}, pages 6522--6530.

\bibitem[{Weng et~al.(2025)Weng, Zhu, Bao, Zhang, Wang, Zhang, and Yang}]{weng2024cycleresearcher}
Yixuan Weng, Minjun Zhu, Guangsheng Bao, Hongbo Zhang, Jindong Wang, Yue Zhang, and Linyi Yang. 2025.
\newblock \href {https://openreview.net/forum?id=bjcsVLoHYs} {Cycle{R}esearcher: {I}mproving automated research via automated review}.
\newblock In \emph{The Thirteenth International Conference on Learning Representations}.

\bibitem[{Wu et~al.(2021)Wu, Ribeiro, Heer, and Weld}]{wu-etal-2021-polyjuice}
Tongshuang Wu, Marco~Tulio Ribeiro, Jeffrey Heer, and Daniel Weld. 2021.
\newblock \href {https://doi.org/10.18653/v1/2021.acl-long.523} {Polyjuice: Generating counterfactuals for explaining, evaluating, and improving models}.
\newblock In \emph{Proceedings of the 59th Annual Meeting of the Association for Computational Linguistics and the 11th International Joint Conference on Natural Language Processing (Volume 1: Long Papers)}, pages 6707--6723, Online. Association for Computational Linguistics.

\bibitem[{Wuehrl et~al.(2024)Wuehrl, Wright, Klinger, and Augenstein}]{wuehrl-etal-2024-understanding}
Amelie Wuehrl, Dustin Wright, Roman Klinger, and Isabelle Augenstein. 2024.
\newblock \href {https://doi.org/10.18653/v1/2024.findings-acl.369} {Understanding fine-grained distortions in reports of scientific findings}.
\newblock In \emph{Findings of the Association for Computational Linguistics: ACL 2024}, pages 6175--6191, Bangkok, Thailand. Association for Computational Linguistics.

\bibitem[{Xu et~al.(2025)Xu, Lu, Schoenebeck, and Kong}]{xu2024benchmarking}
Shengwei Xu, Yuxuan Lu, Grant Schoenebeck, and Yuqing Kong. 2025.
\newblock \href {https://openreview.net/forum?id=uE84MGbKD7} {Benchmarking {LLM}s' judgments with no gold standard}.
\newblock In \emph{The Thirteenth International Conference on Learning Representations, {ICLR} 2025, Singapore, April 24-28, 2025}. OpenReview.net.

\bibitem[{Yu et~al.(2024)Yu, Ding, Tan, Luo, Weng, Gong, Zeng, Cui, Han, Sun, Wu, Lan, and Li}]{yu-etal-2024-automated}
Jianxiang Yu, Zichen Ding, Jiaqi Tan, Kangyang Luo, Zhenmin Weng, Chenghua Gong, Long Zeng, RenJing Cui, Chengcheng Han, Qiushi Sun, Zhiyong Wu, Yunshi Lan, and Xiang Li. 2024.
\newblock \href {https://doi.org/10.18653/v1/2024.findings-emnlp.595} {Automated peer reviewing in paper {SEA}: Standardization, evaluation, and analysis}.
\newblock In \emph{Findings of the Association for Computational Linguistics: EMNLP 2024}, pages 10164--10184, Miami, Florida, USA. Association for Computational Linguistics.

\bibitem[{Yuan et~al.(2022)Yuan, Liu, and Neubig}]{yuan2022can}
Weizhe Yuan, Pengfei Liu, and Graham Neubig. 2022.
\newblock Can we automate scientific reviewing?
\newblock \emph{Journal of Artificial Intelligence Research}, 75:171--212.

\bibitem[{Zhang and Abernethy(2025)}]{zhang2025reviewing}
Tianmai~M Zhang and Neil~F Abernethy. 2025.
\newblock \href {https://arxiv.org/abs/2505.23824v2} {Reviewing scientific papers for critical problems with reasoning {LLM}s: {B}aseline approaches and automatic evaluation}.
\newblock \emph{arXiv preprint arXiv:2505.23824v2}.

\bibitem[{Zheng et~al.(2023)Zheng, Chiang, Sheng, Zhuang, Wu, Zhuang, Lin, Li, Li, Xing, Zhang, Gonzalez, and Stoica}]{zheng2023judging}
Lianmin Zheng, Wei-Lin Chiang, Ying Sheng, Siyuan Zhuang, Zhanghao Wu, Yonghao Zhuang, Zi~Lin, Zhuohan Li, Dacheng Li, Eric Xing, Hao Zhang, Joseph~E Gonzalez, and Ion Stoica. 2023.
\newblock \href {https://proceedings.neurips.cc/paper_files/paper/2023/file/91f18a1287b398d378ef22505bf41832-Paper-Datasets_and_Benchmarks.pdf} {Judging {LLM}-as-a-judge with {MT}-bench and chatbot arena}.
\newblock In \emph{Advances in Neural Information Processing Systems}, volume~36, pages 46595--46623. Curran Associates, Inc.

\bibitem[{Zhou et~al.(2023)Zhou, Muresanu, Han, Paster, Pitis, Chan, and Ba}]{zhou2023large}
Yongchao Zhou, Andrei~Ioan Muresanu, Ziwen Han, Keiran Paster, Silviu Pitis, Harris Chan, and Jimmy Ba. 2023.
\newblock \href {https://openreview.net/forum?id=92gvk82DE-} {Large language models are human-level prompt engineers}.
\newblock In \emph{The Eleventh International Conference on Learning Representations}.

\bibitem[{Zhu et~al.(2025)Zhu, Weng, Yang, and Zhang}]{zhu2025deepreview}
Minjun Zhu, Yixuan Weng, Linyi Yang, and Yue Zhang. 2025.
\newblock \href {https://arxiv.org/abs/2503.08569v1} {Deepreview: {I}mproving {LLM}-based paper review with human-like deep thinking process}.
\newblock \emph{arXiv preprint arXiv:2503.08569v1}.

\end{thebibliography}
\bibliographystyle{acl_natbib}

\newpage

\appendix

\section{Dataset Preprocessing} \label{asec:preproc}
\subsection{Underlying Dataset Preprocessing}
Since the counterfactual generation aims to make surgical edits of few tokens, the accuracy of parsing of papers needs to be high with the tables being particularly challenging. We therefore cross-match the accepted papers with ar5iv\footnote{\url{https://ar5iv.labs.arxiv.org/}(last accessed on 20/07/25)} providing HTML versions of the papers constructed from the LaTeX source using title and author for retrieval. Roughly one quarter of the papers across all conferences can be matched exactly. We convert those original papers to markdown using their HTML version as a reference. Subsequently, we filter out papers for which the title or abstract could not be identified and where the number of sections lies below three suggesting an error. 

\subsection{Research Logic Extraction} \label{assec:rle}
\subsubsection{Pseudo-code}
We report the pseudo-code for extracting the research logic of a scientific paper in Algorithm \autoref{alg:rl}. Due to space restrictions we defer the reader to the acompanying code of this paper for the detailed prompts used during each individual step.

\begin{algorithm*}
\begin{algorithmic}
\REQUIRE Input = title, abstract, text
\ENSURE Output = \fin, \con, \res, \met, supports, anchors, coreferences

\STATE RG $\leftarrow$ llm\_identify\_research\_goal(title, abstract, text)
\STATE CON,$\text{anchors}_\fin$ $\leftarrow$ llm\_extract\_contributions(title, abstract, text)
\STATE CON $\leftarrow$ llm\_rank\_contributions($\con$, RG)
\STATE $\fin \leftarrow$ $\{c \in \text{CON} | c \text{ is findings}\}$
\STATE \;
\STATE $\con, \text{anchors}_\con \leftarrow$ llm\_extract\_conclusions(abstract, text)
\STATE supports$_\fin$ $\leftarrow$ llm\_link\_conclusions\_to\_findings($\con$, $\fin$, title, abstract)
\STATE \;
\STATE $\res, \text{anchors}_\res \leftarrow$ llm\_extract\_results(abstract, text)
\STATE supports$_\con$ $\leftarrow$ llm\_link\_results\_to\_conclusions($\res$, $\con$, title, abstract)
\STATE \;
\STATE $\met \text{anchors}_\met \leftarrow$ llm\_extract\_methods(abstract, text)
\STATE supports$_\res$ $\leftarrow$ llm\_link\_method\_to\_results($\met$, $\res$, title, abstract)
\STATE \;
\STATE coreferences $\leftarrow$ llm\_extract\_coreferences(abstract, text, $\fin$, $\con$, $\res$, $\met$)
\STATE \;
\RETURN $\fin, \con, \res, \met, \bigcup_{i\in\{\fin, \con, \res\}} \text{supports}_i, \bigcup_{i\in\{\fin, \con, \res\}} \text{anchors}_i, \text{coreferences}$
\end{algorithmic}
\caption{Pseudo-code for research logic extraction. Each function starting with 'llm' uses LLMs in zero-shot mode with a single self-refinement step.}
\label{alg:rl}
\end{algorithm*}

\subsubsection{Hyper-parameters}

For the research logic extraction we query the OpenAI API\footnote{\url{https://openai.com/} (last accessed on 20/07/25)} between May and June 2025. We use GPT-4o-mini in version \texttt{2024-07-18}. For generation, we use the default hyper-parameters; i.e. a temperature of 1.0 and no nucleus sampling.

\subsection{Counterfactual Generation} \label{assec:cfgen}

\subsubsection{Prompts}

The generation of counterfactuals involves multiple steps and prompts including self-refinement. Due to space restrictions, we report on the exact prompts used for counterfactual generation in the supplementary code. \autoref{fig:general_prompt} summarizes the general prompt structure which is common to all steps in the pipeline.

\begin{figure}\centering\begin{minipage}{0.95\linewidth}
\centering
  \prompttemplate{%
  [system persona prompt] \\

[brief summary of the task] \\

\#\#\# DEFINITIONS 
[definition of terms and concepts] \\

\#\#\# INPUTS 
[structured list of inputs] \\

\#\#\# SPECIFIC INSTRUCTIONS 
[step-by-step instructions] \\

\#\#\# OUTPUT FORMAT 
[description of output JSON format] \\

\#\#\# IMPORTANT POINTS 
[reiterating hard constraints and requirements] \\
  }
  \caption{General prompt design used during dataset creation.}\label{fig:general_prompt}
\end{minipage}
\end{figure}

\subsubsection{Hyper-parameters}

\paragraph{Soundness-critical Counterfactuals}
For the soundness-critical counterfactuals, we use GPT-4o-mini with the same configuration as before; i.e. version \texttt{2024-07-18} with a temperature of 1.0 queried in May and June 2025.

\paragraph{Soundness-neutral Counterfactuals}
For the soundness-neutral counterfactuals, we use Phi-4 14B run with temperature of 0.8 and top-40 sampling. We run the ollama version.\footnote{\url{https://ollama.com/library/phi4} (last accessed on 20/07/2025)}

\section{Experiments} \label{asec:experiments}

\subsection{ARG Zeroshot Prompts}

\autoref{fig:genericarg} shows the prompt used for the \textsc{Zero-Generic} approach. \autoref{fig:specificarg} shows the prompt used for the \textsc{Zero-Guide} approach. The key difference, is a brief description of the reviewing approach provided as a template parameter based on publicly available reviewer guidelines of each venue.

\begin{figure*}\centering\begin{minipage}{0.95\linewidth}
\centering
  \prompttemplate{%
  You are an expert peer reviewer with extensive experience in academic publishing. \\
  Your goal is to provide a rigorous, objective, and constructive review strictly adhering to the provided template. \\
  You are reviewing for \textbf{\{venue\}}. \textbf{\{venue\}} has the following focus: \textbf{\{venue\_description\}} \\

\#\# Input Paper \\
\textbf{\{paper\}} \\

\#\# Reviewer Instructions \\
* Read the entire paper carefully and critically \\
* Systematically evaluate the paper using the provided template \\
* Ensure each section of the template is completed precisely and substantively \\
* Ground your assessment in specific evidence from the paper \\
* Maintain an objective, professional tone \\

\#\# Output Format \\
\-\-\-\-\- \\
Review Report \\
\textbf{\{template\}} \\

\#\# Evaluation Approach \\
* Analyze the paper comprehensively across multiple dimensions \\
* Provide specific, actionable feedback \\
* Support all assessments with direct references to the paper \\
* Be precise in addressing each required template section \\
* Critically assess the paper's scientific merit and contribution \\

\#\# Final Reminder: \\
* Strictly follow the output structure \\
* Provide substantive content for each template section \\
* Ensure your scores align with your qualitative assessment \\
* For each score, pick only numbers from the respective range and by its meaning explained above \\
* Make sure to start your review report with ``\-\-\-\-\-''
  }
  \caption{Generic ARG prompt.}\label{fig:genericarg}
\end{minipage}
\end{figure*}

\begin{figure*}\centering\begin{minipage}{0.95\linewidth}
\centering
  \prompttemplate{%
   You are an expert peer reviewer with extensive experience in academic publishing. \\
    Your goal is to provide a rigorous, objective, and constructive review strictly adhering to the provided template. \\
    You are reviewing for \textbf{\{venue\}}. \textbf{\{venue\}} has the following focus: \textbf{\{venue\_description\}} \\

    \#\# Input Paper \\
   \textbf{\{paper\}} \\

    \#\# Reviewing Guidelines \\
    You should closely follow the guidelines of the conference provided below while reviewing. \\

    \textbf{\{venue\_guidelines\}} \\

    \#\# Output Format \\
    \-\-\-\-\- \\
    Review Report \\
    \textbf{\{template\}} \\

    \#\# Evaluation Approach \\
    * Analyze the paper comprehensively across multiple dimensions \\
    * Provide specific, actionable feedback \\
    * Support all assessments with direct references to the paper \\
    * Be precise in addressing each required template section \\
    * Critically assess the paper's scientific merit and contribution \\

    \#\# Final Reminder: \\
    * Strictly follow the output format \\
    * Provide substantive content for each template section \\
    * Closely follow the review guidelines \\
    * Ensure your scores align with your qualitative assessment \\
    * For each score, pick only numbers from the respective range and by its meaning explained above \\
    * Make sure to start your review report with ``\-\-\-\-\-''
   }
  \caption{Guided ARG with detailed instructions for reviewing chosen for each venue in the dataset.} \label{fig:specificarg}
\end{minipage}
\end{figure*}

\subsection{ARG Hyper-parameters} \label{asec:hyper}
For all ARGs we discard the paper's appendix since this frequently lead to errors due to the limited context window of the models. If a paper exceeds the context window of an LLM according to the recommended ranges of best performance \cite{hsieh2024ruler}, we truncate the paper at the end.

For \textsc{Zero-Generic-GPT4om}, \textsc{Zero-Guide-GPT4om}, MARG, TreeReviewer, and for fall-back parsing of the fine-tuned ARG outputs we use GPT-4o-mini version \texttt{2024-07-18} with a temperature of 0 queried in May to July 2025. For \textsc{Zero-Generic-GPT4.1} we use version \texttt{gpt-4\_1-2025-04-14} with a temperature of 0 queried in July 2025. For the DeepSeekV3-based zero-shot ARGs we query the DeepSeek API\footnote{\url{https://api-docs.deepseek.com}} using version \textsc{DeepSeek-V3-0324} as a chat model with temperature 0 queried in July 2025.  We adapt the publicly available code of MARG\footnote{\url{https://github.com/allenai/marg-reviewer/tree/master} (last accessed on 18.07.25)} to use GPT-4o-mini. Likewise, we adapt the public code of TreeReviewer\footnote{\url{https://github.com/YuanChang98/tree-review}} to use GPT-4o-mini instead of the original use of Gemini-2.0-flash as in the original paper to ensure a better comparability of approaches irrespective of the underlying LLM.

For the fine-tuned models Reviewer2 and DeepReviewer, we use the default hyperparameters provided by the authors; we do not set the temperature to 0 since this resulted in frequent output errors (repeated tokens etc.) that could not be parsed into a review report. Reviewer2 uses two fine-tuned models\footnote{\url{https://huggingface.co/GitBag/Reviewer2_Mr} and \url{https://huggingface.co/GitBag/Reviewer2_Mp} (last accessed on 18.07.25)} based on Llama-2-7B-Chat \cite{llama2} with a temperature of $0.6$. DeepReviewer is based on Phi-4-14B \cite{abdin2024phi}\footnote{\url{https://huggingface.co/WestlakeNLP/DeepReviewer-14B}} with a temperature of $0.4$.

\section{Complementary Results}

\subsection{Detailed Statistical Testing Results} \label{assec:stat_details}

\autoref{tab:pvalues} reports the detailed statistical test results.

\begin{table*}[h] 
\centering
\footnotesize
\begin{tabular}{r c c c c c c}
\toprule
& \multicolumn{2}{c}{\textbf{Aspects}} & \multicolumn{2}{c}{\textbf{Sentiment}} & \multicolumn{2}{c}{\textbf{Score}} \\
\cmidrule(lr){2-3} \cmidrule(lr){4-5} \cmidrule(lr){6-7}
& p-value & corrected & p-value & corrected & p-value & corrected \\
\midrule
\textsc{oracle} & $1.3 \cdot 10^{-17}$ & $\mathbf{1.6 \cdot 10^{-16}}$ & $1.6 \cdot 10^{-112}$ & $\mathbf{2.1 \cdot 10^{-111}}$ & $0.0$ & $\mathbf{0.0}$\\
\midrule
Reviewer2 & $0.641$ & $0.842$ & $0.488$ & $0.705$ & $0.145$ & $0.420$ \\
DeepReviewer & $0.712$ & $0.842$ & $0.559$ & $0.727$ & $0.162$ & $0.420$ \\
\midrule
\textsc{Zero-Guide-DeepSeek14B} & $0.466$ & $0.842$ & $0.488$ & $0.705$ & $0.208$ & $0.450$ \\
\textsc{Zero-Generic-DeepSeek14B} & $0.938$ & $0.938$ & $0.011$ & $0.068$ & $0.463$ & $0.670$ \\
\textsc{Zero-Guide-DeepSeekV3} & $0.513$ & $0.842$ & $0.116$ & $0.303$ & $0.805$ & $0.805$ \\
\textsc{Zero-Generic-DeepSeekV3} &$0.813$ & $0.880$ & $0.936$ & $0.936$ & $0.604$ & $0.714$ \\
\textsc{Zero-Guide-Phi4} & $0.617$ & $0.842$ & $0.786$ & $0.852$ & $0.515$ & $0.670$ \\
\textsc{Zero-Generic-Phi4} & $0.503$ & $0.842$ & $0.035$ & $0.153$ & $0.505$ & $0.670$ \\
\textsc{Zero-Guide-GPT4om} & $0.503$ & $0.842$ & $0.161$ & $0.303$ & $0.431$ & $0.670$ \\
\textsc{Zero-Generic-GPT4om} & $0.144$ & $0.842$ & $0.654$ & $0.773$ & $0.096$ & $0.415$ \\
\textsc{Zero-Generic-GPT4.1} & $0.242$ & $0.842$ & $0.163$ & $0.303$ & $0.058$ & $0.380$ \\
\midrule
\textsc{TreeReviewer} & $0.687$ & $0.842$ & $0.121$ & $0.303$ & $0.772$ & $0.805$ \\
\bottomrule
\end{tabular}
\caption{For all ARGs, the estimated p-values (uncorrected and BH-corrected for multiple testing per delta) by the linear mixed effect model. The significant corrected p-values for $\alpha = 0.05$ are in bold face.}
\label{tab:pvalues}
\end{table*}

\subsection{Sensitivity to Paper Surface Form}

\autoref{tab:surface} reports the ROUGE-2 and Jaccard overlap of the assertions in the pairs of original and counterfactual reviews.

\begin{table*}[h] 
\centering
\small
\begin{tabular}{r c c}
\toprule
\textbf{ARG} & \textbf{ROUGE-2} & \textbf{Assertion Jaccard} \\
\midrule
\textsc{oracle} & $0.57\pm.05$ & $0.96\pm.10$\\
\midrule
Reviewer2 & $0.38\pm.34$ & $0.32\pm.20$ \\
DeepReviewer & $0.25\pm.10$ & $0.27\pm.13$\\
\midrule
\textsc{Zero-Guide-DeepSeek14B} & $0.26\pm.07$ & $0.39\pm.16$\\
\textsc{Zero-Generic-DeepSeek14B} & $0.27\pm.05$ & $0.42\pm.17$ \\
\textsc{Zero-Guide-DeepSeekV3} & $0.43\pm0.09$ & $0.51\pm0.19$ \\
\textsc{Zero-Generic-DeepSeekV3} & $0.39\pm0.09$ & $0.49\pm0.17$ \\

\textsc{Zero-Guide-GPT4om} & $0.54\pm.10$ & $0.55\pm.20$ \\
\textsc{Zero-Generic-GPT4om} & $0.54\pm.11$ & $0.57\pm.19$ \\

\textsc{Zero-Guide-Phi4} & $0.37\pm.11$ & $0.44\pm.18$\\
\textsc{Zero-Generic-Phi4} & $0.39\pm.11$ & $0.48\pm.18$ \\
\bottomrule
\end{tabular}
\caption{The ROUGE-2 and overlap of assertions in terms of the Jaccard index for all pairs reviews for the original and soundness-neutral-edited papers. For both metrics, the higher the more similar the two reviews.}
\label{tab:surface}
\end{table*}

\end{document}